\documentclass[10pt]{article}

\usepackage{arxiv}
\usepackage{natbib}

\usepackage[T2A,T1]{fontenc}%
\usepackage[utf8]{inputenc}%
\usepackage{textcomp}%
\usepackage[russian,french,english]{babel}%
\usepackage{linguex}%
\usepackage{array}
\usepackage{graphicx}

\usepackage{hyperref}

\title{GLeMM: A large-scale multilingual dataset for morphological research}
\renewcommand{\shorttitle}{\textsc{GLeMM}}
\author{%
  \mbox{%
    \href{https://orcid.org/0000-0003-4492-171X}{\includegraphics[scale=0.06]{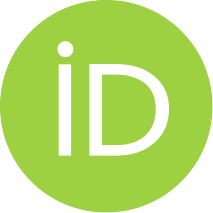}\hspace{1mm}Nabil Hathout}%
  }\\
	CLLE, CNRS, Université Toulouse Jean Jaurès, Toulouse, France\\
	\texttt{nabil.hathout@univ-tlse2.fr}\\
	\And
  \mbox{%
  \href{https://orcid.org//0000-0002-0160-7512}{\includegraphics[scale=0.06]{orcid.pdf}\hspace{1mm}Basilio Calderone}%
  }\\
	CLLE Montaigne, CNRS, Université Bordeaux Montaigne, Pessac, France\\
	\texttt{basilio.calderone@u-bordeaux-montaigne.fr} \\
	\And
  \mbox{%
  \href{https://orcid.org//0000-0002-6144-3011}{\includegraphics[scale=0.06]{orcid.pdf}\hspace{1mm}Fiammetta Namer}%
  }\\
	ATILF, Université de Lorraine, CNRS, Nancy, France\\
	\texttt{fiammetta.namer@univ-lorraine.fr} \\
	\And
  \mbox{%
  \href{https://orcid.org/0000-0001-9439-3658}{\includegraphics[scale=0.06]{orcid.pdf}\hspace{1mm}Franck Sajous}%
  }\\
	CLLE, CNRS, Université Toulouse Jean Jaurès, Toulouse, France\\
	\texttt{franck.sajous@univ-tlse2.fr}
}

\begin{document}

\maketitle

\begin{abstract}
  In derivational morphology, what mechanisms govern the variation in form-meaning relations between words?
The answers to this type of questions are typically based on intuition and on observations drawn from limited data, even when a wide range of languages is considered.
Many of these studies are difficult to replicate and generalize.
To address this issue, we present GLeMM, a new derivational resource designed for experimentation and data-driven description in morphology. 
GLeMM is characterized by (\emph{i}) its large size, (\emph{ii}) its extensive coverage (currently amounting to seven European languages, i.e., German, English, Spanish, French, Italian, Polish, Russian, (\emph{iii}) its fully automated design, identical across all languages, (\emph{iv}) the automatic annotation of morphological features on each entry, as well as (\emph{v}) the encoding of semantic descriptions for a significant subset of these entries.
It enables researchers to address difficult questions, such as the role of form and meaning in word-formation, and to develop and experimentally test computational methods that identify the structures of derivational morphology.
The article describes how GLeMM is created using Wiktionary articles and presents  various case studies illustrating possible applications of the resource.  
\end{abstract}

\section{Introduction}
\label{sec:introduction}

In derivational morphology, what mechanisms govern the variation in form-meaning relations between words?
For example, why do speakers form \emph{scarcity}\textsubscript{N} from \emph{scarce}\textsubscript{ADJ} but \emph{loneliness}\textsubscript{N} from \emph{lonely}\textsubscript{ADJ}, even though both derivatives express the same relation to their base?
Are there similar differences in all languages?
And conversely, are some mechanisms specific to certain language groups?
The answers to these questions are typically based on intuition and on observations drawn from limited data, even when a wide range of languages is considered.
Many of these studies are difficult to replicate and generalize. They are not based on corpora or public datasets and they lack tools capable of characterizing and measuring key phenomena of derivational morphology, such as discrepancies between meaning and form.

To address this issue, we present a new derivational resource designed for experimentation and data-driven description in morphology. 
This resource, named GLeMM\footnote{%
  This article only covers version 2.0 of GLeMM.
  It is available under a CC-BY-4.0 license on the git repository: \url{https://src.koda.cnrs.fr/nabil.hathout.1/glemm}.
  Two earlier versions (1.3 and 1.4) are also available on the same site.%
} (short for ``Gros Lexique Morphologique Multilingue''\footnote{%
  \emph{Big Morphological Multilingual Lexicon}.  GLeMM is more precisely a multilingual collection of comparable derivational lexicons.%
}) provides formal and semantic descriptions of derivational relations.

GLeMM is characterized by several remarquable features : (\emph{i}) its large size, (\emph{ii}) its extensive coverage (currently amounting to seven European languages, i.e., German, English, Spanish, French, Italian, Polish, Russian, (\emph{iii})  its fully automated design, identical across all languages, (\emph{iv}) the automatic annotation of morphological features on each entry, as well as (\emph{v}) the encoding of semantic descriptions for a significant subset of these entries.
This resource is designed to help morphologists bridge the gap between derivational morphology and inflectional morphology which offers more detailed and comprehensive descriptions, based on more advanced theories and covering a larger number of languages.
In particular, GLeMM enables researchers to address difficult questions, such as the role of form and meaning in word-formation (WF), and to develop and experimentally test computational methods that identify the structures of derivational morphology.

The remainder of the article is organized as follows. Section~\ref{sec:rationale} outlines the rationale behind the study and the development of GLeMM. Section~\ref{sec:sota} reviews some of the existing  resources for derivational morphology.
Section~\ref{sec:data} and~\ref{sec:method} describe how GLeMM is created using Wiktionary articles.  We then present the resource itself (Section~\ref{sec:results}) and various case studies illustrating possible applications (Section~\ref{sec:use-cases}). Section~\ref{sec:conclusion} outlines future avenues of research and offers a brief conclusion.

\section{Rationale}
\label{sec:rationale}

General-purpose morphological resources such as CELEX \citep{baayen1995.celex} are often used to create materials for psycholinguistic experiments, such as that of \cite{dejong2000.family-size} on the effect of morphological family size or that of \cite{hay2003.phonotactics} on speech perception.
They are also employed in studies based on text corpora, as in \citeauthor{plag2009.suffix_ordering}'s (\citeyear{plag2009.suffix_ordering}) study on affix ordering. 
By contrast, they are not used as ``lexical corpora'' for the study and description of numerous topics in derivational morphology, as those addressed at major conferences like the \emph{International Morphology Meeting}: constraint violation, marginal word-formation, derivational paradigms, affix polysemy and homonymy, and meaning-form discrepancy.
Examples of phenomena related to these topics include affix rivalry, conversion, back-formation, parasynthetic constructions, suppletion, and allomorphy.
Similar questions that come to mind include the  affixation / compounding (unclear) boundaries, the type of graph formed by derivational relations, or the role played by form and meaning in word-formation.
There are two main reasons why it is difficult to use existing general-purpose morphological resources for this type of study:
\begin{enumerate}
\item Their coverage is insufficient to adequately reflect the diversity of morphologically complex lexemes.
 This type of use requires very large datasets with sufficiently varied entries to include instances of atypical phenomena, which actually are significantly rarer than typical phenomena.
\item They do not provide a descriptions of derivational  relations  and constructions at the three levels traditionally considered in morphology: form, category, and meaning.
\end{enumerate}

One of GLeMM's goals is to overcome these limitations and contribute to a better understanding of these phenomena. In Section~\ref{sec:use-cases}, we present several case studies that illustrate the possibilities offered by this new resource.

\section{Derivational resources}
\label{sec:sota}

Derivational resources are characterized by how their design, by the properties they describe (e.g., semantic annotations), and by their intended purpose.

\subsection{Reference resources created or validated by humans}
\label{sec:sota-human}

Derivational resources can be distinguished according to whether or not they have undergone systematic manual annotation or revision.
This is the case for only a few of them.
For the seven languages we are interested in, these include Derivatario \citep{talamo2016.derivatario}, CELEX \citep{baayen1995.celex}, CatVar \citep{habash2003.catvar}, and Démonette\footnote{%
  A small portion of the entries in Démonette-2.0 come from resources created using programs, specifically DériF \citep{namer2009.hermes} and Morphonette \citep{hathout2011.lel}.
  Their entries haven't been manually checked.
  The other sources used to feed Démonette (Convers \citep{tribout2010.phd,tribout2012.word-structure}, Dénom \citep{strnadova2014.phd}, DiMoc \citep{roche2004.verbum,%
    plenat2005.breves-remarques-ette,%
    roche2011.linguistica,%
    lignon2011.dumal,%
    plenat2014.foisonnement-at,%
    roche2016.cmlf%
  }, Lexeur \citep{wauquier2020.zwjw}, Mordan \citep{koehl2012.phd,koehl2014.morphology}, VerbAction \citep{tanguy2002.taln}) were all created manually, and their entries can be used in gold standards of French derivational morphology.%
  
} \citep{hathout2014.lilt,hathout2016.lrec-demonette,namer2023.demonette}.
Because they have been manually annotated or revised, these resources can be considered as gold standards.
On the other hand, they are generally labor-intensive and limited in scope.
Therefore, they are suitable for creating experimental materials for psycholinguistic studies, but are often too small for data-driven analysis and for training statistical and neural networks.
  
\subsection{Resources created (semi-)automatically}
\label{sec:sota-smi-automatic}

Resources created (semi-)automatically are more numerous. These include the DeriNet lexicons for Czech \citep{sevcikova2014.LREC,derinet2019.database}, Russian \citep{kyjanek2022.derinet-RU}, Polish and Spanish \citep{lango2018.derinet-ES-PL,rn1925}, as well as DerivBase for German \citep{zeller2013.derivbase,zeller2014.coling} and Russian \citep{vodolazsky2020.derivbase-RU}.
These resources were created using rules that implement morphological analyses derived from reference works and textbooks.
The rules are applied to lemmas present in tagged and lemmatized corpora.
The produced candidates are then post-edited to filter out incorrect analyses.
A comprehensive revision of the DeriNet and DerivBase corpora is not feasible given the cost of such a review.
However, several of these resources have been evaluated using manually annotated or revised samples. 
Precision, as reported in \citep[p.\ 1856]{lango2018.derinet-ES-PL}, \citep{zeller2013.derivbase} and \citep{vodolazsky2020.derivbase-RU}, ranges from 85\% for DeriNet.ES (Spanish), to 93\% for DerivBase-1.4 (German), and 95\% for DeriNet.RU (Russian).
Overall, this level of accuracy is too low for these resources to be considered gold standards.

Universal Derivation (UDer; \citealt{kyjanek2020.uder,uder-data}) covers 21 languages.
It incorporates the resources listed above and reorganizes them into rooted trees.
UniMorph-4.0 \citep{batsuren2022.unimorph-short} is a similar resource that contains derivational lexicons for 30 languages.
Several UniMorph lexicons are taken from MorphyNet \citep{batsuren2021.morphynet}, a resource that provides derivational lexicons for 15 languages: Finnish, Serbo-Croatian, Italian, Hungarian, Russian, Spanish, French, Portuguese, Polish, German, Czech, English, Catalan, Swedish, and Mongolian.
MorphyNet is created by parsing the etymological sections of Wiktionary and then supplementing the lexicons with lexeme pairs identified from the relations existing between their cognates.
The cognates are taken from  CogNet\footnote{%
CogNet is a database containing 8.1 million pairs of cognates from 338 languages.
It was created using data from WordNet and Wiktionary.%
} \citep{batsuren2019.cognet}.

One takeaway from this very limited overview is that the vast majority of morphological derivational resources have not been manually verified and that some contain a significant number of errors (for a more comprehensive inventory, see \citealt{kyjanek2018.resources}).

\subsection{Semantics}
\label{sec:sota-semantics}

Resources containing semantic annotations are relatively rare.
For English, \citet{fellbaum2009.morphosemantic-wn3} created an English derivational resource consisting of lexeme pairs extracted from Princeton WordNet \citep{miller1990.wordnet,fellbaum98}.
where the derived words are annotated with the semantic role they play with respect to their base (\emph{agent}, \emph{material}, \emph{event}, \emph{location}, \emph{instrument}, etc.).
In WordNet, the meanings of lexemes are part of synsets, and synsets have definitions.
However, the semantic relations between morphologically complex lexemes and their bases cannot be retrieved from  these definitions or elsewhere in WordNet.
Other resources include similar semantic annotations. For example, a subset of the Czech DeriNet entries includes semantic descriptions of the complex lexemes using five of \citeauthor{bagasheva2017.semantic-concepts}'s (\citeyear{bagasheva2017.semantic-concepts}) semantic categories: diminutive, possessive, feminine noun, iterative, and aspectual meaning  \citep{sevcikova2019.semantics-derinet}.
For French, ontological annotation has been proposed for a subset of the entries in Démonette-2.0 \citep{rn1870} using tags adapted from Fr-SemCore \citep{rn1661}.

To our knowledge, the only resource with semantic annotations comparable to those found in GLeMM is Démonette-1.2 \citep{hathout2014.lilt}, where entries are equipped with semantic patterns that define the target lexemes with respect to the meaning of the source lexemes.

\section{Data used to create GLeMM}
\label{sec:data}

The primary goal of creating GLeMM is to develop derivational resources that contain as few erroneous (i.e., not morphologically related) pairs of lexemes as possible and whose annotations are as accurate as possible.
This can only be achieved if the development is based on semantic descriptions.
Experience suggests that such descriptions can be heterogeneous and coarse-grained.
For example, \cite{hathout2001.taln,hathout2002.lrec.wordnet} uses synonymy relations to create derivational resources based on DicoSyn for French and WordNet \citep{miller1990.wordnet,fellbaum1999.wordnet} for English. 
Later, \cite{hathout2008.textgraphs3,hathout2009.taln,hathout2009.decembrettes,hathout2011.lel,hathout2014.ls} proposed a method in which semantic proximity is determined by the shared words between definitions from the TLFi \citep{dendien2003.tal}. 
Along the same lines, \cite{hathout2014.lg-lp} explored the possibility of using subgraphs from the dependency analyses of GLAWI definitions \citep{sajous2015.glawi,hathout2016.lrec-glawi}.

For GLeMM, we assume that if a word appears in the definition of some headword, then it is very likely that their meanings are related. 
If this semantic connection is combined with a formal connection, there is a strong likelihood that the word included in the definition is morphologically related to the headword. 
 In this case, the definition corresponds to what \cite{martin1992.robert.pls} calls a \emph{derivational definition}.
This assumption amounts to viewing derivational definitions as explicit clues to lexical motivation.
When a lexeme is used in this way to define another one, their semantic relation is likely to reflect a morphological dependency within the same word family. The method used to create GLeMM  operationalizes the notion of morphological motivation and applies it as a large-scale criterion.

\subsection{Derivational definitions}
\label{sec:derivational-definitions}

Derivational definitions are definitions where the meaning of the defined term is described in relation to that of another lexeme which it is morphologically related to. This is the case in the definitions in \ref{exe:morphological-definitions} taken from Wiktionary where the  morphologically related lexeme  is in bold.

\ex.\label{exe:morphological-definitions}
\a.\label{exe:spryness-pair} spryness = The property of being \textbf{spry}.
\b.	unastounding = Not \textbf{astounding}.
\c.	concretization = The process of \textbf{concretizing} a general principle or idea by delineating, particularizing, or exemplifying it.
\d.	contractualization = The process of \textbf{contractualizing}.
\e.	unmendably = Such that it cannot be \textbf{mended}; in a way that is beyond repair.
\f.	cancerization = Transformation into a \textbf{cancerous} form

The definitions in \ref{exe:morphological-definitions} describe the semantic relation between the defined word and another word in its family.
 The derivational relation can be direct (a single word-formation step, as in \emph{concretize}\textsubscript{V} $\rightarrow$ \emph{concretization}\textsubscript{N}), indirect with multiple word-formation steps, as in \emph{mend}\textsubscript{V} $\rightarrow$ \emph{unmendably}\textsubscript{ADV}, or even more complex, involving a path composed of both ascending and descending steps, as in \emph{cancerous}\textsubscript{A} $\rightarrow$ \emph{cancerization}\textsubscript{N}, which links two descendants of \emph{cancer}\textsubscript{N}. 
In all these examples, the definitions show how the meaning of the defined term can be derived from the source word.

One obvious obstacle to using this type of definition is that dictionaries do not explicitly labeled them as such, any more than they label other types of definitions.
There is no way of knowing \emph{a priori} whether a definition is morphological or not, especially since we do not know which pairs of lexemes are morphologically related and which are not; the answer to this second question is precisely the aim of our work. 
The FAPinette method presented in Section~\ref{sec:method} offers a solution to this problem. It is based on the identification of formal analogies  that exist between the defined words and the words that appear in their definitions.

\subsection{Wiktionary definitions}
\label{sec:wiktionanry-definitions}

GLeMM uses the derivational definitions from Wiktionary.
This dictionary project offers many advantages.
It is free.
It is available in a wide range of languages.
The editions in the various languages are structured in a similar way: one page per form.
On the flip side, Wiktionary articles are written in wikicode, a markup language whose syntax is not formally defined and which does not require explicit markup for all the objects that make up the articles.
This makes it difficult to use.
Fortunately, there are versions of Wiktionary that have already been parsed, which we use.
These versions include  GLAWI  for French \citep{sajous2015.glawi,hathout2016.lrec-glawi},  GLAW-IT  for Italian \citep{calderone2016.sle}, and  ENGLAWI  for English \citep{sajous2020.lrec-englawi}. 
These three dictionaries are in XML and use a similar set of tags for all three languages.
The parsing of definitions, morphological sections, and information on form and category is accurate and reliable.
However, the lack of wikicode markup of the information contained in the etymological sections prevents its use in creating GLeMM lexicons. 
We also use the dictionaries from the Kaikki project \citep{ylonen2022.lrec}.
Like the previous ones, these dictionaries are parsed versions of Wiktionary editions.
Kaikki covers a larger and constantly growing number of languages.
In addition, its dictionaries are built using recent dumps of Wiktionary. 
When we launched the development of GLeMM in 2024, Kaikki did not contain a dictionary for Italian.
It did, however, contain dictionaries for the six other languages covered by GLeMM (see Table~\ref{tab:glemm-source-lexicons}).

\subsection{Morphological sections}
\label{sec:morphological-sections}

Wiktionary articles include sections that provide a list of words morphologically related to headwords.
The nature and quality of the information they contain vary to some extent.
For example, in the article for the English verb \emph{interpret}, the ``Derived terms'' section contains the lexemes listed under \ref{exe:interpret-derives-terms}, and the ``Related terms'' section contains the lexemes listed under \ref{exe:relatex-terms}.

\ex.\label{exe:interpret-derives-terms}
interpretory, overinterpret, preinterpret

\ex.\label{exe:relatex-terms}
interpretable,  interpretation, interpretative, interpret away, interpreter, interpretive, misinterpret, reinterpret

In the GLAWI dictionaries (a term we use to refer to the GLAWI, GAW-IT, and ENGLAWI dictionaries), the relevant sections are labelled ``derivatives'', ``compounds'', and ``related words''.
In the Kaikki dictionaries, these sections are  named``derived'' and ``related''. 
The relations contained in these sections were used to create GLeMM because they broaden the coverage of its lexicons and they strengthen the formal regularities which FAPinette is based on.
However, they is not entirely reliable, as suggested by the somewhat random distribution of derivatives in the ``Derived terms'' \ref{exe:interpret-derives-terms} and ``Related terms'' \ref{exe:relatex-terms} sections.
Consequently, the lexeme pairs extracted from these sections undergo the same selection processes as the ones extracted from the definitions (see Sections~\ref{sec:analogy-signature} and~\ref{sec:alternation-patterns}).

\subsection{MorphyNet}
\label{sec:data-morphynet}

We also added entries from the MorphyNet lexicons to the candidates considered for the identification of formal analogies. The lexeme pairs extracted from MorphyNet are symmetrized (i.e., made bidirectional).  However, unlike the pairs from the morphological sections, those from MorphyNet are reliable enough and therefore are all retained in the GLeMM lexicons.

\subsection{Summary}
\label{sec:data-summary}

Table~\ref{tab:glemm-source-lexicons} lists the sources used to produce the candidate pairs for each of the languages covered by GLeMM.
\begin{table}[th]
  \centering
  \caption{Source of the candidate pairs used to create the GLeMM lexicons}
  \label{tab:glemm-source-lexicons}
  \begin{tabular}{llllll}
    \hline
& \multicolumn{2}{c}{\textbf{GLAWI}}    & \multicolumn{2}{c}{\textbf{Kaikki}} & \textbf{MorphyNet} \\
&  \textbf{definitions} & \textbf{morphological} &  \textbf{definitions} & \textbf{morphological} & \textbf{entries} \\
&              &  \textbf{section}      &             &  \textbf{section}      &  \\
\hline
\textbf{English} & yes & yes & yes & yes & yes \\
\textbf{French} & yes & yes & yes & yes & yes \\
\textbf{Italian} & yes & yes &  &  & yes \\
\textbf{German} &  &  & yes & yes & yes \\
\textbf{Polish} &  &  & yes & yes & yes \\
\textbf{Spanish} &  &  & yes & yes & yes \\
\textbf{Russian} &  &  & yes & yes & yes \\
\hline
  \end{tabular}
\end{table}

\section{Method}
\label{sec:method}

GLeMM largely follows the method used by \cite{hathout2020.lrec-glawinette} to create the French derivational lexicon Glawinette, but differs in several respects: in the number of languages it covers, in its use of a broader set of initial resources, and in improvements made to the FAPinette method these authors designed to select pairs of morphologically related lexemes and determine their exponents.

\subsection{Formal analogy}
\label{sec:formal-analogy}

Identifying morphologically related words from a set of candidate word pairs is undoubtedly the most challenging problem to solve when creating a resource such as GLeMM.
For example, how can we figure out that the noun \emph{spryness}\textsubscript{N} and the adjective \emph{spry}\textsubscript{ADJ} in \ref{exe:spryness-pair} are morphologically related?
How can this relation be distinguished from the ones that connect \emph{spryness}\textsubscript{N} to the other words in its definition (e.g., \emph{property}\textsubscript{N}, \emph{be}\textsubscript{V})?
In FAPinette, this identification is based on formal analogy \citep{lepage1998.coling,lepage2003.hdr,lepage2004.analogy-formal-languages,stroppa2005.conll,stroppa2005.these,langlais2008.coling,lavallee2009.morphochallenge}
A formal analogy A:B::C:D is a relationship between four strings, $A$, $B$, $C$, and $D$, that can be aligned such that the differences between $A$ and $B$ are identical to those between $C$ and $D$.
For example, \texttt{abbc}:\texttt{bbd}::\texttt{aefc}:\texttt{fed} forms an analogy in that the differences between the forms in the first pair are that \texttt{a} is deleted and \texttt{c} is replaced by \texttt{d}, as seen in the alignment in \ref{exe:formal-analogy-alignment-1}, where these differences are the same as those between the two forms in the second pair \ref{exe:formal-analogy-alignment-2}.
\ex.\label{exe:formal-analogy-alignment-1-2}
\a.\label{exe:formal-analogy-alignment-1}%
\begin{tabular}[t]{|l|l|l|}
  \hline
  \textbf{a}&bb&\textbf{c}\\
  \hline
  &bb&\textbf{d}\\
  \hline
\end{tabular}
\b.\label{exe:formal-analogy-alignment-2}%
\begin{tabular}[t]{|l|l|l|}
  \hline
  \textbf{a}&ef&\textbf{c}\\
  \hline
  &ef&\textbf{d}\\
  \hline
\end{tabular}

Returning to example \ref{exe:spryness-pair}, the English edition of Wiktionary includes derivational definitions for several nouns denoting properties ending in \emph{\mbox{-ness}}, as in \ref{exe:derivatives-ness}.
\ex.\label{exe:derivatives-ness}
\a.\label{exe:derivatives-abruptness}	abruptness = The state of being abrupt or broken.
\b.\label{exe:derivatives-circularness}	circularness = The state or quality of being circular.
\c.\label{exe:derivatives-suspectfulness}	suspectfulness = The quality of being suspectful.
\d.\label{exe:derivatives-tightness}	tightness = The quality or degree of being tight

If we compare the form \emph{spryness}\textsubscript{N} with the words \emph{property}\textsubscript{N} and \emph{spry}\textsubscript{ADJ} that appear in its definition, the pair \emph{spryness}\textsubscript{N} / \emph{property\textsubscript{N}} exhibits the differences shown in \ref{exe:spryness-property-alignment}: the deletion of the initial \texttt{s}, the insertion of the sequence \texttt{opert}, and the deletion of the final sequence \texttt{ness}.
These differences do not appear in any other pair consisting of a Wiktionary headword and another word from its definition.
It does not correspond to any regularity and can therefore be ignored.
More specifically, the lack of regularity is a strong indication of a lack of derivational relation between the two words.
\ex.\label{exe:spryness-property-alignment}%
\begin{tabular}[t]{|l|l|l|l|l|}
  \hline
  \textbf{s}&pr& & y & \textbf{ness}\\
  \hline
  &pr&\textbf{opert} &y & \\
  \hline
\end{tabular}

When \emph{spry}\textsubscript{ADJ} is aligned with \emph{spryness}\textsubscript{N} \ref{exe:spryness-spry-alignment}, their difference is the presence of \texttt{ness} at the end of the noun.
The same difference occurs between the pairs \emph{abruptness}\textsubscript{N} and \emph{abrupt}\textsubscript{ADJ} \ref{exe:abruptness-abrupt-alignment}, which can be derived from \ref{exe:derivatives-abruptness}, \emph{circularness}\textsubscript{N} and \emph{circular}\textsubscript{ADJ} from \ref{exe:derivatives-circularness}, \emph{suspectfulness}\textsubscript{N} and \emph{suspectful}\textsubscript{ADJ} from \ref{exe:derivatives-suspectfulness}, and so on. 
In other words, \emph{spry}\textsubscript{ADJ} / \emph{spryness}\textsubscript{N} forms analogies with a significant number of other word pairs, suggesting that the difference is systematic and likely indicates the existence of a derivational morphological relation between these word pairs.
\ex.
\a.\label{exe:spryness-spry-alignment}%
\begin{tabular}[t]{|l|l|}
  \hline
  spry&\textbf{ness}\\
  \hline
  spry&\\
  \hline
\end{tabular}
\b.\label{exe:abruptness-abrupt-alignment}%
\begin{tabular}[t]{|l|l|}
  \hline
  abrupt&\textbf{ness}\\
  \hline
  abrupt&\\
  \hline
\end{tabular}

The FAPinette method makes use of another property of formal analogies, namely that if $A$:$B$::$C$:$D$, then $A$:$C$::$B$:$D$ (referred to as the ``exchange of the means'' by \citealt{lepage2003.hdr}). 
For example, the existence of an analogy \texttt{abbc}:\texttt{bbd}::\texttt{aefc}:\texttt{fed} entails the existence of an analogy \texttt{abbc}:\texttt{aefc}::\texttt{bbd}:\texttt{fed} \ref{exe:formal-analogy-alignment-3-4} in which the differences correspond to the identical sub-sequences in \ref{exe:formal-analogy-alignment-1-2}.
\ex.\label{exe:formal-analogy-alignment-3-4}
\a.\label{exe:formal-analogy-alignment-3}%
\begin{tabular}[t]{|l|l|l|}
  \hline
  a&\textbf{bb}&c\\
  \hline
  a&\textbf{ef}&c\\
  \hline
\end{tabular}
\b.\label{exe:formal-analogy-alignment-4}%
\begin{tabular}[t]{|l|l|}
  \hline
  \textbf{bb}&d\\
  \hline
  \textbf{ef}&d\\
  \hline
\end{tabular}

Once identified, the word pairs belonging to the same analogy system such as \ref{exe:formal-analogy-alignment-3-4}  can be used to determine the sub-sequences that may correspond to regularities.
These sub-sequences may correspond to derivational affixes (e.g., \emph{\mbox{-ness}}), or inflectional markers (see first example  in Table~\ref{tab:spanish-pairs-eamples}), i.e., to overt exponents of the derivational relation connecting the word pair members.

\subsection{Analogy signature}
\label{sec:analogy-signature}

The implementation of FAPinette relies on the ability to quickly identify pairs that can form analogies.
The identification method is approximate but extremely effective.
It associates a signature with each word pair, such that two pairs with identical signatures may form an analogy, whereas  two pairs with different signatures do not.
These signatures, developed by \cite{lepage1998.coling}, consist of the Levenshtein edit distance, and the differences, for each character in the alphabet, between the number of its occurrences in the two words.
For example, for a pair like \emph{spryness}\textsubscript{N} / \emph{spry}\textsubscript{ADJ}, the edit distance is $4$, and the differences are $-1$ for \texttt{n}, $-1$ for \texttt{e}, and $-2$ for \texttt{s}. This signature is shared by all the pairs with which \emph{spryness}\textsubscript{N} / \emph{spry}\textsubscript{ADJ} forms analogies.
By contrast, for the pair \emph{spryness}\textsubscript{N} / \emph{property}, the edit distance is $10$ and the differences are \texttt{n}: $-1$, \texttt{o}: $+1$, \texttt{p}: $+1$, \texttt{r}: $+1$, \texttt{s}: $-3$, \texttt{t}: $+1$.
Furthermore, this pair is the only one with this signature and therefore cannot form an analogy with any other pair. 
A word pair that has the same signature as a large enough number of other pairs likely share formal regularities with the latter which indicates that its words are morphologically related.

The first step in FAPinette consists of (\emph{i}) forming candidate word pairs made up of a Wiktionary headword and a word from both its definition and morphological sections, (\emph{ii})  adding MorphyNet entries  as candidates, (\emph{iii}) computing the analogical signatures of all these pairs, and (\emph{iv}) retaining only the word pairs with signatures  shared by at least 4 other ones (i.e., having a minimal frequency of 5).

\subsection{Alternation patterns}
\label{sec:alternation-patterns}

The next step  associates each word pair with a set of alternation patterns that describe the differences between the two words and, consequently, the transformations that result in the production of the second word from the first.
For example, in the \emph{spry}\textsubscript{ADJ} / \emph{spryness}\textsubscript{N} pair, the final substring \texttt{ness} is present in the second word but not in the first.
The difference reflects how linguists describe this morphological relation, namely as a suffixation where the (orthographic) exponent is \emph{\mbox{-ness}}.
It can be formally described using a pattern composed of a pair of regular expressions such as \texttt{\^{}(.+)ness\$} / \texttt{\^{}(.+)\$}, where\texttt{ (.+)} represents  an identical substring in both words, namely the stem \texttt{spry}\textsubscript{ADJ}.

In some pairs, such as \emph{positivist}\textsubscript{N} / \emph{positivism}\textsubscript{N}, the minimal alternation \texttt{\^{}(.+)t\$} /  \texttt{\^{}(.+)m\$} differs from the way morphologists would describe the relation, that is, as an indirect or cross-formation relation involving the alternation between suffixes \emph{\mbox{-ist}} and \emph{\mbox{-ism}}. 
This more linguistically relevant difference can be identified by collecting all pairs of words that share the same signature as \emph{positivist}\textsubscript{N} and \emph{positivism}\textsubscript{N}, such as \emph{feminist}\textsubscript{N} and \emph{feminism}\textsubscript{N}, \emph{structuralist}\textsubscript{N} and \emph{structuralism}, etc., and forming all their analogies through exchange of the means. 
These analogies highlight that the lexemes on the left (\emph{positivist}\textsubscript{N}, \emph{feminist}\textsubscript{N}, \emph{structuralist}\textsubscript{N}, etc.) share a final sub-sequence \texttt{ist}, and that the lexemes on the right (\emph{positivism}\textsubscript{N}, \emph{feminism}\textsubscript{N}, \emph{structuralism}\textsubscript{N}, etc.) share a final sub-sequence \texttt{ism}, thus yielding an alternation pattern  \texttt{\^{}(.+)ist\$} /  \texttt{\^{}(.+)ism\$} which is more consistent with the linguistic description of the morphological relation.

At the end of this second stage, a second round of selection is conducted, which consists of retaining only those candidate pairs whose analogies yield sufficiently general alternations, i.e., alternations that applies to a minimum number of word pairs.

\subsection{FAP selection}
\label{sec:fap-selection}

The third step in FAPinette is the selection of a FAP (for \emph{fine-grained alternation pattern}) for each pair, that is, the pattern which best describes the morphological relation between the two words.
The various analogies in which a pair occurs generally yield several patterns. Table~\ref{tab:alternation-patterns-positivist-positivism} presents some of the patterns obtained from analogies between \emph{positivist}\textsubscript{N} and \emph{positivism}\textsubscript{N} and other candidate pairs.
\begin{table}[th]
  \centering
    \caption{Examples of alternation patterns yielded by analogies involving the pair \emph{positivist}\textsubscript{N} and \emph{positivism}\textsubscript{N}}
  \label{tab:alternation-patterns-positivist-positivism}
  \begin{tabular}{llllll}
    \hline
\textbf{A} & \textbf{B} & \textbf{C} & \textbf{D} &\textbf{ left pattern} & \textbf{right pattern}\\
    \hline
positivist & positivism & feminist & feminism & \texttt{\^{}(.+)ist\$} & \texttt{\^{}(.+)ism\$}\\
positivist & positivism & feminist & feminism & \texttt{\^{}(.+)i(.+)ist\$} & \texttt{\^{}(.+)i(.+)ism\$}\\
positivist & positivism & activist & activism & \texttt{\^{}(.+)itivist\$} & \texttt{\^{}(.+)tivism\$}\\
positivist & positivism & activist & activism & \texttt{\^{}(.+)tivist\$} & \texttt{\^{}(.+)tivism\$}\\
positivist & positivism & activist & activism & \texttt{\^{}(.+)ivist\$} & \texttt{\^{}(.+)ivism\$}\\
positivist & positivism & activist & activism & \texttt{\^{}(.+)vist\$} & \texttt{\^{}(.+)vism\$}\\
positivist & positivism & activist & activism & \texttt{\^{}(.+)ist\$} & \texttt{\^{}(.+)ism\$}\\
positivist & positivism & catastrophist & catastrophism & \texttt{\^{}(.+)s(.+)ist\$} & \texttt{\^{}(.+)s(.+)ism\$}\\
positivist & positivism & catastrophist & catastrophism & \texttt{\^{}(.+)ist\$} & \texttt{\^{}(.+)ism\$}\\
    \hline
  \end{tabular}
\end{table}

Among these patterns, \texttt{\^{}(.+)ist\$} / \texttt{\^{}(.+)ism\$} is distinguished by the fact that it consists of expressions that connect the largest number of pairs.
This connectivity stems from the alternation of \emph{\mbox{-ist}} (i.e., \texttt{\^{}(.+)ist\$}) with \emph{\mbox{-y}} (i.e., \texttt{\^{}(.+)y\$}) in pairs such as \emph{biologist}\textsubscript{N} / \emph{biology}\textsubscript{N}, with \emph{\mbox{-ics}} in pairs such as \emph{pragmatist}\textsubscript{N} / \emph{pragmatics}\textsubscript{N}, with \emph{\mbox{-ize}} in pairs such as \emph{apologist}\textsubscript{N} / \emph{apologize}\textsubscript{V}, and so on. The same is true for \emph{\mbox{-ism}} (i.e., \texttt{\^{}(.+)ism\$}).
By contrast, in the \texttt{\^{}(.+)tivist\$} / \texttt{\^{}(.+)tivism\$} pattern, the expression  \texttt{\^{}(.+)tivist\$} appears in patterns for far fewer word pairs namely the ones containing words ending in \texttt{itivist} such as \emph{activist}\textsubscript{N}, \emph{collectivist}\textsubscript{N}, \emph{nativist}\textsubscript{N}, etc.
 The same goes for \texttt{\^{}(.+)tivism\$} . In other words,  \texttt{\^{}(.+)ist\$} and \texttt{\^{}(.+)ism\$} can be used, either together or separately, to describe a greater number of pairs than  \texttt{\^{}(.+)tivist\$} and \texttt{\^{}(.+)tivism\$}.

For each pair, FAPinette selects the pattern composed of the most connecting expressions, that is, those shared by the greatest number of words appearing in pairs selected at the end of the first two processing steps.
The fact that many pairs share an exponent suggests that it represents a stable, formal regularity within the lexicon.
Selecting the most closely related patterns amounts to favoring alternations that correspond to the most general (i.e., linguistically sound) derivational schemas.

\subsection{Comparison of FAPinette with other morphological analysis methods}
\label{sec:comparison-fapinette-other-methods}

The description of morphological relations using rules or patterns is common in inflectional morphology. 
One of the classic methods is \citeauthor{albright2003.mgl}'s (\citeyear{albright2003.mgl}) minimal generalization learner (MGL), which generalizes the alternations between pairs of phonemic forms.
Generalization is based upon the phonological features of alternating phonemes and the features of phonemes in their contexts. 
More recently, \cite{bonami2016.ws,beniamine2018.phd,beniamine2021.SCIL} proposed the Qumin analyzer, which generalizes alternation patterns and their contexts into inflectional classes.
A key difference between FAPinette, MGL and Qumin is that the two latter tools identify alternations at the word pair level.
FAPinette is more complex and operates in three stages: (\emph{i}) selection of subsets of candidate pairs that are likely to be morphologically related;  (\emph{ii}) identification of the formal patterns of the lexemes in each subset; (\emph{iii}) alignment of the patterns.
In this way, FAPinette performs two tasks: computing formal characterizations of the derivational relations between morphologically related lexemes and filtering out pairs whose alternations are not sufficiently regular (i.e., frequent). 
This second function is essential for processing the derivation.
In contrast, tools designed for inflection do not filter out erroneous form pairs because they are all assumed to be correct.

Candidate selection is typically performed by morphological analyzers that process inflection and derivation simultaneously.
One example is Linguistica \citep{goldsmith2001.cl,goldsmith2006.nle}, which parses the forms in a raw corpus.
The parser splits the forms into possible stems and affixes.  It then identifies the segments that allow the words in the corpus to be reconstructed while minimizing the length of their description.
Similarly, Morfessor \citep{creutz2002.acl,creutz2004.sigorphon,creutz2005.morfessor} combines four machine learning methods that identify pairs of morphologically related words and segment them into morphemes.
Unlike Linguistica, Morfessor can segment words into more than two morphemes.
More recently, \cite{cotterell2018.joint} proposed a method that combines morphological segmentation into a stem and affixes with an estimation of semantic coherence based on word embeddings.
This type of tokenization remains relevant today: LLMs typically segment corpora and prompts into tokens using algorithms such as byte-pair encoding (BPE; \citealt{gage1994.BPE}).

\section{Results}
\label{sec:results}

\subsection{Languages}
\label{sec:results-languages}

GLeMM contains lexicons for seven European languages: three Romance languages (French, Italian and Spanish), two Germanic languages (German and English), and two Slavic languages (Polish and Russian). This selection was based on the resources available at the start of GLeMM's development.

The resource allows users to study the same phenomena in languages whose derivational morphology exhibits various degrees of diversity.
For example, in French and German, word formation differs in terms of trade-offs between derivation and compounding.
French tends to favor affixation (e.g., \emph{poissonnier}\textsubscript{N} `fisherman'), whereas German seems to prefer compounding (e.g., \emph{Fischhändler}\textsubscript{N} `fisherman'). 
Another example is the perfective  / imperfective opposition which distinguishes Slavic languages from other language families.  In the former, perfective and imperfective verbs are regarded as separate lexemes (e.g., \emph{budować}\textsubscript{V} `build.\textsc{ipfv}'  \emph{vs} \emph{zbudować}\textsubscript{V} `build.\textsc{pfv}' in Polish).
This significant morphological hypothesis, discussed in particular by \cite{sevcikova2025.habilitation}, has major implications for the derivational system of these languages, particularly regarding the treatment of conversion, which is neither entirely inflectional nor entirely derivational \citep{hledikova2024.conversion}.

\subsection{Data structure}
\label{sec:data-structure}

GLeMM  entries consist of pairs of morphologically related lexemes belonging to the major grammatical categories: noun (N), verb (V), adjective (A), and adverb (R). The lexicon for each language is made up of two tables: \texttt{pairs.tsv} and \texttt{defs.tsv}. 

\subsubsection{Table \texttt{pairs.tsv}}
\label{sec:results-table-pairs-tsv}

Entries in \texttt{pairs.tsv} describe the morphological relation of lexeme pairs.  Lexemes are represented by their lemmas. More specifically, entries include the lemmas of the related lexemes (lemma1, lemma2), their parts-of-speech (cat1, cat2), their morphological  pattern (exponent1, exponent2), and a shared sub-sequence (stem).
Table~\ref{tab:spanish-pairs-eamples} shows three entries from the Spanish \texttt{pairs.tsv} table.  As we already indicated, the morphological patterns, or exponents, are described using regular expressions, in which the expressions \texttt{(.+)} correspond to a substring shared by lemma1 and lemma2, that is, the value of stem.  The exponents describe the differences between the lemmas of the lexemes in the pair.
For example, for the pair \emph{ducha}\textsubscript{N} / \emph{duchita}\textsubscript{N}, exponent2 indicates that \emph{duchita}\textsubscript{N} is a lexeme suffixed with \emph{\mbox{-ita}}; the expression \texttt{(.+)} corresponds to the orthographic sequence \texttt{duch}. Note that the exponent can be null, as in the case of \emph{talento}\textsubscript{N} in the last two pairs in the table.
\begin{table}[th]
  \centering
  \caption{Examples of entries in the Spanish \texttt{pairs.tsv} table}
  \label{tab:spanish-pairs-eamples}
  \begin{tabular}{lllllll}
    \hline
\textbf{lemma1} & \textbf{cat1} & \textbf{lemma2} & \textbf{cat2} & \textbf{stem} & \textbf{exponent1} & \textbf{exponent2}\\
    \hline
rastrero & A & rastreramente & R & rastrer & \texttt{\^{}(.+)o\$} & \texttt{\^{}(.+)amente\$}\\
`despicable.M' &  & `despicably' &  &  &  & \\
    \hline
ducha & N & duchita & N & duch & \texttt{\^{}(.+)a\$} & \texttt{\^{}(.+)ita\$}\\
`shower' &  & `small shower' &  &  &  & \\
    \hline
talentoso & A & talento & N & talento & \texttt{\^{}(.+)so\$} & \texttt{\^{}(.+)\$}\\
`talented.M' &  & `talent' &  &  &  & \\
    \hline
talento & N & talentoso & A & talento & \texttt{\^{}(.+)\$} & \texttt{\^{}(.+)so\$}\\
    \hline
  \end{tabular}
\end{table}

The first two examples in Table~\ref{tab:spanish-pairs-eamples} illustrate ``classic'' word-formations: the suffixation in \emph{\mbox{-mento}}, which is used to form adjective-based adverbs, and the suffixation in \emph{\mbox{-ita}}, which is used to form diminutives. The last two examples illustrate the symmetry of the relations in this table.

Table~\ref{tab:english-pairs-eamples} presents examples from the English \texttt{pairs.tsv} table that show the diversity of derivational relations described in the GLeMM lexicons.
The first pair relates a neoclassical compound to its hyponym \citep{bauer2017.compounding}.  The left compounding element \emph{\mbox{electro-}} is provided by the exponent2 pattern.
In the second example, \emph{clonally}\textsubscript{ADV} / \emph{subclonally}\textsubscript{ADV} is an indirect derivational relation, which  complements the \emph{subclonal}\textsubscript{ADJ} / \emph{subclonally}\textsubscript{ADV} direct, regular one which also occurs in the \texttt{pairs.tsv} table.
Finally, the third pair describes a complex relationship involving a prefixation in \emph{\mbox{super-}} and a suffixation in \emph{\mbox{-ical}}. 
\begin{table}[th]
  \centering
  \caption{Examples of entries in the English \texttt{pairs.tsv} table}
  \label{tab:english-pairs-eamples}
  \begin{tabular}{lllllll}
    \hline
\textbf{lemma1} & \textbf{cat1} & \textbf{lemma2} & \textbf{cat2} & \textbf{stem} & \textbf{exponent1} & \textbf{exponent2}\\
    \hline
sensitivity & N & electrosensitivity & N & sensitivity & \texttt{\^{}(.+)\$} & \texttt{\^{}electro(.+)\$}\\
    \hline
clonally & R & subclonally & R & clonally & \texttt{\^{}(.+)\$} & \texttt{\^{}sub(.+)\$}\\
    \hline
supertechnological & A & technology & N & technolog & \texttt{\^{}super(.+)ical\$} & \texttt{\^{}(.+)y\$}\\
    \hline
  \end{tabular}
\end{table}

\subsubsection{Table \texttt{defs.tsv}}
\label{sec:results-table-defs-tsv}

Table \texttt{defs.tsv} contains a subset of the entries from \texttt{pairs.tsv} for which Wiktionary provides a definition of the second lexeme that includes the first one.
As a consequence, the second lexeme is expected to be derived from first.
Entries in \texttt{defs.tsv} provide the same information as in \texttt{pairs.tsv} (the lemmas and categories of both lexemes, a derivational stem, and two exponents), plus a definition of the second lexeme that contains the first.
The definition is given in its original form and in lemmatized form. Tables~\ref{tab:english-defs-eamples} and~\ref{tab:french-defs-eamples} give examples of entries from the \texttt{defs.tsv} table in English and French.

\begin{table}[th]
  \centering
  \caption{Examples of entries in the English \texttt{defs.tsv} table}
  \label{tab:english-defs-eamples}
  \begin{tabular}{ll>{\raggedright\arraybackslash}p{0.65\textwidth}}
    \hline
a. &  & spry	A	spryness	N	spry	\texttt{\^{}(.+)\$}	\texttt{\^{}(.+)ness\$}\\
& \textbf{definition2} & The property of being spry.\\
 & \textbf{lemmatized\_definition2} & the property of be spry .\\
    \hline
b. &  & astounding	A	unastounding	A	astounding	\texttt{\^{}(.+)\$}	\texttt{\^{}un(.+)\$}\\
 & \textbf{definition2} & Not astounding.\\
 & \textbf{lemmatized\_definition2} & not astounding .\\
    \hline
c. &  & concretize	V	concretization 	N	concretiz	\texttt{\^{}(.+)e\$}		\texttt{\^{}(.+)ation\$}\\
 & definition2 & The process of concretizing a general principle or idea by delineating, particularizing, or exemplifying it.\\
 & \textbf{lemmatized\_definition2} & the process of concretize a general principle or idea by delineate , particularize , or exemplify it .\\
    \hline
d. &  & contractualize	V	contractualization	N	contractualiz	\texttt{\^{}(.+)e\$}	\texttt{\^{}(.+)ation\$}\\
 & \textbf{definition2} & The process of contractualizing.\\
 & \textbf{lemmatized\_definition2} & the process of contractualize .\\
    \hline
    e. &  & approximation	N	approximative	A	approximat	\texttt{\^{}(.+)ion\$}	\texttt{\^{}(.+)ive\$}\\
 & \textbf{definition2} & Of, relating to, or being an estimate or approximation.\\
 & \textbf{lemmatized\_definition2} & of , relate to , or be a estimate or approximation .\\
    \hline
f. &  & mend	V	unmendably	R	mend	\texttt{\^{}(.+)\$}	\texttt{\^{}un(.+)ably\$}\\
 & \textbf{definition2} & Such that it cannot be mended; in a way that is beyond repair.\\
 & \textbf{lemmatized\_definition2} & such that it can not be mend ; in a way that be beyond repair .\\
    \hline
  \end{tabular}
\end{table}

In each  example, the first line reproduces the content of \texttt{pairs.tsv}.
The next two lines provide the definition of the second lexeme in its original (definition2) and lemmatized forms (lemmatized\_definition2).
For the same word-formation process, the expression of the semantic relation is also variable, as can be seen in the definitions of \emph{concretization}\textsubscript{N} (c) and \emph{contractualization}\textsubscript{N} (d).
Example (e) illustrates the fact that semantics do not always matches  morphological orientation: although \emph{approximative}\textsubscript{ADJ} is defined with respect to \emph{approximation}\textsubscript{N}, none of the two words derives from the other.
More generally, some pairs in \texttt{defs.tsv} are not in a base / derivative relation, as in (f), where the relation corresponds to a double affixation (\emph{\mbox{-able}} $+$ \emph{\mbox{-ly}}).

\begin{table}[th]
  \centering
  \caption{Examples of entries in the French \texttt{defs.tsv} table}
  \label{tab:french-defs-eamples}
  \begin{tabular}{ll>{\raggedright\arraybackslash}p{0.65\textwidth}}
    \hline
    a. &  & coffrage 	N	décoffrer 	V	coffr	\texttt{\^{}(.+)age\$}	\texttt{\^{}dé(.+)er\$} \\
    & & `formwork'	`unmould' \\
 & \textbf{definition2} & Retirer le coffrage après prise du béton. \\
 &  & `Remove the formwork after the concrete has set.' \\
 & \textbf{lemmatized\_definition2} & retirer le coffrage après prise de le béton . \\
    \hline
    b. &  & neutralité	N	neutraliser	V	neutral	\texttt{\^{}(.+)ité\$}	\texttt{\^{}(.+)iser\$} \\
     & &  `neutrality' `neutralize' \\
 & \textbf{definition2} & Rendre un pays, une ville, un navire neutre, assurer à un pays, à une ville, à un navire l'état de neutralité. \\
 &  & `To make a country, city, or ship neutral; to ensure that a country, city, or ship has the status of neutrality.' \\
 & \textbf{lemmatized\_definition2} & rendre un pays , un ville , un navire neutre , assurer à un pays , à un ville , à un navire le état de neutralité . \\
    \hline
    c. &  & infirmerie 	N	infirmier	N	infirm	\texttt{\^{}(.+)erie\$}	\texttt{\^{}(.+)ier\$} \\
    & & `infirmary' `nurse.Mxs'\\
 & \textbf{definition2} & Celui qui soigne et sert les malades dans une infirmerie, dans un hôpital, dans une ambulance. \\
 &  & `Someone who cares for and serves sick people in a medical facility, hospital, or ambulance.' \\
       & \textbf{lemmatized\_definition2} & celui qui soigner et servir le malade dans un infirmerie , dans un hôpital , dans un ambulance . \\
    \hline
d. &  & particule 	N	interparticulaire	A	particul	\texttt{\^{}(.+)e\$}	\texttt{\^{}inter(.+)aire\$} \\
     & & `particule'  `interparticular' \\
 & \textbf{definition2} & Qui relie les particules. \\
 &  & `Which connects particles.' \\
 & \textbf{lemmatized\_definition2} & qui relier le particule . \\
    \hline
  \end{tabular}
\end{table}

The examples in Table~\ref{tab:french-defs-eamples} illustrate various phenomena of French derivational morphology.
In (a), the stem  of the verb (i.e., \emph{décoffr}) does not match the stem of the base noun \emph{coffrage}\textsubscript{N}, but rather derives from that of another member of the family, i.e., \emph{coffre}\textsubscript{N}. The mechanism is similar to the one proposed by \cite{hathout2011.dumal} to describe parasynthetic constructions.
Example (b) involves the same mechanism, with the difference that the noun-verb relation is indirect: it connects an \emph{\mbox{-iser}} suffixed verb to a \emph{\mbox{-ité}} suffixed noun, both based on the same adjective, namely \emph{neutre}\textsubscript{ADJ} `neuter', as evidenced by the first part of the definition of \emph{neutraliser}\textsubscript{V}. 
In (c), the direction of the relation is the opposite of what is expected. This asymmetry illustrates what \cite{roche2011.isme-dumal} calls a  relation of ``reciprocal motivation''.
Example (d) is a classical parasynthetic construction \citep{hathout2018.hommage-fradin} where the adjectival suffix \emph{\mbox{-aire}} works just as a plain part-of-speech marker.
These examples illustrate how GLeMM can be used to describe various atypical phenomena in derivational morphology.
The orientation of pairs in \texttt{defs.tsv} allows us to empirically observe motivation relationships within families and to analyze the interplay between formal regularities and semantic dependencies.

\subsection{Size of GLeMM lexicons}
\label{sec:results-size-glemm-lexicons}

Table~\ref{tab:lexicon-size} shows the number of ordered pairs and pair sets of lexemes in the lexicons of the seven languages.
It also gives the number of pairs and pair sets for which the resource provides a semantic description (i.e., a definition). 
Overall, GLeMM provides semantic descriptions for nearly half of the pairs it describes.
The ones without  semantic descriptions originate from MorphyNet and from morphological sections.
It highlights significant differences in the size of the lexicons and in the proportion of entries provided with a definition.
While GLeMM facilitates the study of specific phenomena across multiple languages, the uneven coverage of the lexicon across different languages hinders its use for comparing their derivational morphologies.
The relatively small proportion of Italian lexical entries that have a definition is due to the fact that  the lexeme pairs in the Italian \texttt{pairs.tsv} table come mainly from MorphyNet, and to the fact that, as we mentioned earlier, Kaikki did not include an Italian dictionary at the start of GLeMM's development.
\begin{table}[ht]
  \centering
  \caption{Number of pairs of lexemes, with and without semantic descriptions, in the GLeMM lexicons}
  \label{tab:lexicon-size}
  \begin{tabular}{lrrrrrr}
\hline
& \textbf{ordered pairs} & \textbf{defs} & \textbf{ratio} & \textbf{pair sets} & \textbf{defs} & \textbf{ratio}\\
\hline
\textbf{English} & 779,170 & 271,496 & 0.348 & 395,886 & 254,528 & 0.643\\
\textbf{Russian} & 732,633 & 261,713 & 0.357 & 366,352 & 250,425 & 0.684\\
\textbf{French} & 271,343 & 86,526 & 0.319 & 135,679 & 76,686 & 0.565\\
\textbf{Polish} & 196,433 & 35,581 & 0.181 & 98,222 & 31,533 & 0.321\\
\textbf{German} & 136,248 & 42,438 & 0.311 & 68,127 & 39,995 & 0.587\\
\textbf{Italian} & 121,400 & 9,569 & 0.079 & 60,700 & 8,843 & 0.146\\
\textbf{Spanish} & 70,531 & 19,548 & 0.277 & 35,286 & 17,155 & 0.486\\
\hline
\textbf{Total} & 2,307,758 & 726,871 & 0.268 & 1,160,252 & 679,165 & 0.490\\
\hline
  \end{tabular}
\end{table}

\subsection{Validity of pairs and annotations}
\label{sec:results-validity-pairs-annotatons}

As mentioned above, GLeMM, like many similar resources, has not been manually checked.
As a result, some entries contain errors.
Some pairs in the GLeMM lexicons are not morphologically related. These are often orthographic variants, such as \ref{exe:error-1}. Errors can also result from a strong formal and semantic similarity \ref{exe:error-2} that does not follow any derivational pattern.
Nevertheless, since the acquisition of word pairs is based on derivational definitions and the regularity of alternations, such errors are overall relatively rare (Table~\ref{tab:glemm-evaluation}). They are the flip side of prioritizing coverage and systematicity.
\ex.
\a.\label{exe:error-1}	anonymization / anonymisation; laser / lazer
\b.\label{exe:error-2}	chatter / chitter 

Errors in GLeMM lexicons also regard annotations, specifically alternation patterns. Some of these are suboptimal, as in \ref{exe:apologizer}, where the patterns include the initial \texttt{a}.
As a result, this pair falls into a different derivational series from that of the pair \emph{autonomize} / \emph{autonomy}, which is correctly analyzed.
Furthermore, the error in the alternation patterns induces the prediction of a stem (\texttt{p}) that is, in turn, incorrect.
This suboptimality arises from the fact that the alternation patterns are computed automatically based on formal regularities observed in the data, without any prior ``abstract'' morphological analysis.
\ex.\label{exe:apologizer}
apologizer / apology\\
\texttt{\^{}a(.+)ologizer\$} /  \texttt{\^{}a(.+)ology\$}

To assess the reliability of GLeMM, samples of 100 pairs randomly selected from French, English, and Italian lexicons were manually revised by three of the authors of the article. Table~\ref{tab:glemm-evaluation} shows the number of erroneous pairs and suboptimal patterns, as well as the accuracy for both types of information.
\begin{table}[th]
  \centering
  \caption{Accuracy of the pairs and optimality of the exponents}
  \label{tab:glemm-evaluation}
  \begin{tabular}{lrrrr}
    \hline
& \multicolumn{2}{c}{\textbf{correct}} & \multicolumn{2}{c}{\textbf{optimal}}\\
& \textbf{pairs} & \textbf{accuracy} & \textbf{patterns} & \textbf{accuracy}\\
    \hline
\textbf{English} & 94 & 0.94 & 87 & 0.93\\
\textbf{French} & 99 & 0.99 & 92 & 0.93\\
\textbf{Italian} & 100 & 1.00 & 84 & 0.84\\
    \hline
  \end{tabular}
\end{table}

Overall, GLeMM is a resource that prioritizes lexical coverage and the validity of word pairs over exponent  and stem optimality. 
Its goal is to observe a sufficient number of relations, families, and sub-structures such as triangles \citep{lignon2014.cmlf} or stolons \citep{hathout2025.backformation} to identify principles that determine the morphological organization of the lexicon (see Sections~\ref{sec:use-case-triangles} and~\ref{sec:use-case-stolons}).
Since the errors in the resource do not compromise these observations, we consider them acceptable. For a detailed study of a specific phenomenon, users may need to manually review data extracted from GLeMM. 
Conversely, for more experimental and ``extensive'' studies, the impact of errors on the observations should be relatively minor overall. Furthermore, some of these errors could be identified in the future based on the distributional meaning of lexemes and the semantic representations of their definitions.

\subsection{Comparison with existing resources}
\label{sec:results-comparison-exisiting-resources}

\subsubsection{Size}
\label{sec:results-comparison-size}

GLeMM can be compared to other existing resources in several ways.
Quantitative comparison is the most straightforward, but it can be difficult because derivational resources do not all describe the same objects.
These resources primarily contain three types of description:
\begin{enumerate}
\item descriptions of complex lexemes in the form of a morpheme-based decomposition (group 1);
\item descriptions of derivational relations between lexemes, where the relations  may be limited to direct derivations or may include indirect relations (group 2);
\item descriptions of morphological families (group 3). Families may be structured as graphs, or they may be sets of lexemes.
\end{enumerate}

The first group includes resources such as CELEX \citep{baayen1995.celex} for German, English, and Dutch; Derivatario \citep{talamo2016.derivatario} for Italian; and DeriNet \citep{sevcikova2014.LREC,derinet2019.database} for Czech, Spanish, Polish, and Russian. The quantitative comparison of GLeMM with these resources is based on the number of entries for complex lexemes.
GLeMM belongs to the second group which also includes resources such as Démonette for French \citep{namer2023.demonette} and various multilingual resources such as UDer (\citealt{kyjanek2020.uder,uder-data}, 21 languages), MorphyNet (\citealt{batsuren2021.morphynet}, 15 languages), and UniMorph (\citealt{batsuren2022.unimorph-short}, 30 languages). Since their entries consist of lexeme pairs, the comparison can be based on the number of pairs.
To further refine the comparison by distinguishing, for example, between direct and indirect relationships would require establishing criteria for differentiating between the two types of relations. This task is all the more difficult given that the direct / indirect distinction is primarily theoretical. 
The resources in the third group describes morphological organization using morphological families. This is the case with CatVar \citep{habash2003.catvar} for English, Famorpho-fr \citep{hathout2005.cahlex} and Morphonette \citep{hathout2011.lel} for French, the DerivBase databases \citep{zeller2013.derivbase,zeller2014.coling}  for German, and DerivBase.RU \citep{vodolazsky2020.derivbase-RU}  for Russian. 
We do not compare GLeMM to the first three resources because their families are not structured according to the derivational complexity of the lexemes they contain.  On the other hand, the families in DerivBase are described as ``rooted trees''. For this resource, the comparison is based on the direct (first-level) relations.
Table~\ref{tab:objects-comparison} lists the objects of the resources considered for the  comparison.
For a more detailed and comprehensive overview of existing morphological derivation resources, please refer to \citep{kyjanek2018.resources} and the Universal Derivation website\footnote{%
  \url{https://ufal.mff.cuni.cz/universal-derivations}
  }.
\begin{table}[th]
  \centering
  \caption{Objects of the existing resources considered for the quantitative comparison of GLeMM lexicons}
  \label{tab:objects-comparison}
  \begin{tabular}{lll}
    \hline
\textbf{resources} & \textbf{languages} & \textbf{objects}\\
    \hline
\textbf{CELEX} & German, English, Dutch & complex lexemes\\
\textbf{Démonette} & French & relations (set-pairs)\\
\textbf{Derivatario} & Italian & complex lexemes\\
\textbf{Derinet} & Czech, Russian, Polish, Spanish & complex lexemes\\
\textbf{DerivBase} & German, Russian & first-level relations\\
\textbf{MorphyNet} & 15 languages & relations (set-pairs)\\
\textbf{UDer} & 21 languages & relations (set-pairs)\\
\textbf{UniMorph} & 30 languages & relations (set-pairs)\\
    \hline
  \end{tabular}
\end{table}

Table~\ref{tab:resource-comparison} compares the number of entries in GLeMM (pairs, unordered) with the number of items described in some similar resources, both monolingual and multilingual.
We can see that for the seven languages covered by GLeMM, its lexicons describe more word pairs than those in MorphyNet and UniMorph. 
The same is true for the number of entries in the CELEX corpora.
Only UDer Polish and Spanish lexicons contain more pairs than those in GLeMM. 
Despite these differences, the total number of pair-sets in GLeMM exceeds that of UDer for all seven languages.
The last column of the table shows the number of items described in Démonette-2.0 (F), DerivBase-1.4 (G), and Derivatario (I).
Note that DerivBase-2.0 contains fewer entries than DerivBase-1.4.
Overall, GLeMM contains more entries than comparable resources.
\begin{table}[th]
  \centering
  \caption{Size comparison between GLeMM and other resources}
  \label{tab:resource-comparison}
  \begin{tabular}{lrrrrrrl}
    \hline
\textbf{Language} & \textbf{GLeMM} & \textbf{UDer} & \textbf{MorphyNet} & \textbf{UniMorph} & \textbf{CELEX} & \textbf{Other} & \\
    \hline
\textbf{English} & 395,886 & 32,483 & 225,131 & 225,131 & 52,447 &  & \\
\textbf{Russian} & 366,352 & 303,615 & 93,039 & 14,048 &  &  & \\
\textbf{French} & 135,679 & 13,808 & 72,952 & 73,259 &  & 111,059 & (F)\\
\textbf{Polish} & 98,222 & 214,093 & 58,711 & 58,742 &  &  & \\
\textbf{German} & 68,127 & 43,367 & 29,381 & 29,381 & 51,728 & 63,479 & (G)\\
\textbf{Italian} & 60,700 & 1,783 & 58,848 &  &  & 11,147 & (I)\\
\textbf{Spanish} & 35,286 & 42,825 & 30,777 & 31,293 &  &  & \\
    \hline
\textbf{Total} & 1,160,252 & 651,974 & 568,839 & 431,854 & 104,175\\
    \hline
  \end{tabular}
\end{table}

\subsubsection{Comparison of the WF relations in GLeMM and MorphyNet}
\label{sec:results-comparison-wf-relations}

Not only are GLeMM's pairs more numerous than those in most other existing morphological resources, but they are also more diverse and cover a broader range of derivational morphological phenomena. 
A non-exhaustive comparison of some typical and atypical phenomena described in the English corpora of GLeMM and MorphyNet offers some insight into this.
The comparison is made with MorphyNet because it is arguably the most comparable resource to GLeMM in terms of origin, coverage and description format.
Remember that MorphyNet has been used to build GLeMM, which means that all phenomena that can be observed in MorphyNet are naturally also present in GLeMM. 
The comparison, summarized in Table 3-3, is based on a small number of examples that illustrate some of these phenomena.
No quantification is provided.
\begin{table}[th]
  \centering
  \caption{Comparison of GLeMM (\texttt{pairs.tsv}) and MorphyNet for a few pairs of lexemes whose relations illustrate various derivational configurations.}
  \label{tab:glemm-morphynet-phenomena}
  \begin{tabular}{ll>{\raggedright}p{0.2\textwidth}ll}
    \hline
\textbf{lexeme1} & \textbf{lexeme2} & \textbf{derivational type}  & \textbf{MorphyNet} & \textbf{GLeMM}\\
    \hline
manage\textsubscript{V} & manager\textsubscript{N} & direct derivation & yes & yes \\
measurable\textsubscript{ADJ} & measurability\textsubscript{N} & direct derivation & yes & yes\\
characterize\textsubscript{V} & characterizability\textsubscript{N} & multiple derivations & yes & yes\\
hospitalize\textsubscript{V} & rehospitalization\textsubscript{N} & multiple derivations & no & yes\\
minimalist\textsubscript{N} & minimalism\textsubscript{N} & indirect (cross) relation & no & yes\\
productive\textsubscript{ADJ} & productible\textsubscript{ADJ} & indirect (cross) relation & no & yes\\
walk\textsubscript{V} & walk\textsubscript{N} & conversion & no & no\\
nematode\textsubscript{N} & phytonematode\textsubscript{N} & composition & yes & yes\\
milk\textsubscript{N} & milkman\textsubscript{N} & composition & yes & yes\\
man\textsubscript{N} & milkman\textsubscript{N} & composition & no & yes\\
share\textsubscript{N} & shareholder\textsubscript{N} & composition & no & yes\\
holder\textsubscript{N} & shareholder\textsubscript{N} & composition & no & yes\\
parliament\textsubscript{N} & antiparliamentary\textsubscript{ADJ} & parasynthetic construction & no & yes\\
person\textsubscript{N} & interpersonal\textsubscript{ADJ} & parasynthetic construction & no & yes\\
injury\textsubscript{N} & injure\textsubscript{V} & back-formation & yes & yes\\
burglary\textsubscript{N} & burgle\textsubscript{V} & back-formation & no & yes\\
superficialness\textsubscript{N} & superficiality\textsubscript{N} & overabundance & no & yes\\
adjustation\textsubscript{N} & adjustment\textsubscript{N} & overabundance & no & yes\\
    \hline
  \end{tabular}
\end{table}

As expected, GLeMM and MorphyNet contain pairs of lexemes that are in a direct derivational relation. 
This is the case with the relationship between a verb and its agent noun, such as \emph{manage}\textsubscript{V} $\rightarrow$ \emph{manager}\textsubscript{N}, or between an adjective and its quality noun, such as \emph{measurable}\textsubscript{ADJ} $\rightarrow$ \emph{measurability}\textsubscript{N}. 
Both lexicons also contain examples of relations between a lexeme and a more distant descendant, such as \emph{characterize}\textsubscript{V} $\rightarrow$ \emph{characterizability}\textsubscript{N}, which involves two suffixations. But other pairs, such as \emph{hospitalize}\textsubscript{V} $\rightarrow$ \emph{rehospitalization}\textsubscript{N}, which involve a prefixation and a suffixation, are only present in GLeMM.
The same applies to indirect relations between two derivatives of the same base, such as \emph{minimalist}\textsubscript{N} $\leftrightarrow$ \emph{minimalism}\textsubscript{N} or \emph{productive}\textsubscript{A} $\leftrightarrow$ \emph{productible}\textsubscript{A}, which are present in GLeMM but absent from MorphyNet. 
Note that neither the GLeMM English lexicon nor the MorphyNet English lexicon accounts for conversion because it involves lexemes whose lemmas are identical. For example, a pair such as \emph{walk}\textsubscript{V} $\rightarrow$ \emph{walk}\textsubscript{N} is absent from both lexicons.

While GLeMM and MorphyNet primarily describe derivational relations, they also contain numerous relations that fall under native and neoclassical compounding.
In both resources, compounding may be described by two relations that link each of the compounding elements to the compound. 
For example, both resources contain the relations \emph{milk}\textsubscript{N}$\rightarrow$  \emph{milkman}\textsubscript{N} and \emph{nematode}\textsubscript{N} $\rightarrow$ \emph{phytonematode}\textsubscript{N}. However, only GLeMM also contains the other relation \emph{man}\textsubscript{N} $\rightarrow$ \emph{milkman}\textsubscript{N}. 
Other compounding relations are only found in GLeMM, such as \emph{share}\textsubscript{N} $\rightarrow$ \emph{shareholder}\textsubscript{N} and \emph{holder}\textsubscript{N} $\rightarrow$ \emph{shareholder}\textsubscript{N}.

The structure used in MorphyNet to describe word-formation is similar to the wikicode templates used in the etymological sections of Wiktionary. These templates have two slots: one for a base lexeme and the other for an affix.
This format cannot adequately describe constructions that formally involve a prefix and a suffix, particularly parasynthetic constructions such as \emph{parliament}\textsubscript{N} $\rightarrow$ \emph{antiparliamentary}\textsubscript{ADJ} or \emph{person}\textsubscript{N} $\rightarrow$ \emph{interpersonal}\textsubscript{ADJ}. 
These two pairs are missing in MorphyNet but are included in GLeMM, whose alternation patterns can aggregate multiple exponents.

Back-formation (a phenomenon in which lexeme2 is semantically derived from lexeme1, but its form is simpler than the form of lexeme1) is another phenomenon correctly accounted for by GLeMM.
For example, GLeMM includes the pairs \emph{burglary}\textsubscript{N} $\rightarrow$ \emph{burgle}\textsubscript{V} and \emph{injury}\textsubscript{N} $\rightarrow$ \emph{injure}\textsubscript{V}. 
MorphyNet, on the other hand, only contains the second pair, described as a suffixation in \emph{\mbox{-y}} (\emph{injure}\textsubscript{V} $\rightarrow$ \emph{injury}\textsubscript{N}). 
Overabundance (a configuration in which a base is related to several derivatives derived through rival processes) is another atypical phenomenon that is better described in GLeMM than in MorphyNet.
For example, GLeMM includes pairs such as \emph{superficialness}\textsubscript{N} $\leftrightarrow$ \emph{superficiality}\textsubscript{N} or \emph{adjustation}\textsubscript{N} $\leftrightarrow$ \emph{adjustment}\textsubscript{N}, which are absent in MorphyNet. 

\section{Use cases}
\label{sec:use-cases}

As we mentioned above, GLeMM is designed for experimental linguistics and large-scale corpus studies, which may involve multiple languages. A key feature, shared by other resources, is that the lexicons for all languages are structured identically, and the information provided is of the same type and format. 
For example, GLeMM can be used to carry out a re-word-formation task that adapts the well-known re-inflection task\footnote{
  The reinflection task consists of predicting the inflected form of a lexeme that occurs in a given slot of its inflectional paradigm.%
} to derivation \citep{thyme1994finnish,bonami2016.ws,malouf2017.morphology,cardillo2018.pcfp,kann-schutze-2016-single,silfverberg2018.pcfp,williams2020.form-meaning,guzman2020.analogy} and to test PCFP\footnote{
  The Paradigm Cell Filling Problem (PCFP) is a research question formulated by \cite{ackerman2009.analogy} as: ``What allows for reliable inferences about the inflected (and derived) surface forms of a lexical item?''.%
} \citep{ackerman2009.analogy} in derivation on full lexicons, for example with the aim of identifying the role of form and meaning in WF \citep{hathout2025.form-meaning}.

One strength of GLeMM is that it combines a systematic and relatively detailed description of derivational relations with lexicons with broad and varied coverage, a feature resulting from the fact that they are created from large dictionaries whose primary objective is not morphological description, but rather general lexicographic description. 
Depending on the language, the Wiktionary editions have more entries than those of other dictionaries. This is partly because they include a large number of morphologically complex words.\footnote{
  This is also a consequence of the priority given to the size of the lexicon, its primacy over content quality, and very flexible and inclusive inclusion criteria (rare, obsolete, or specialized words, very recent neologisms, terms from fringe cultures, etc., which rarely find their way into other dictionaries).%
  } As a result, GLeMM can be used as an annotated ``corpus'' from which examples can be collected to describe a range of morphological derivational phenomena.
In the remainder of this section, we present a few examples of phenomena that can be studied using GLeMM. Although we do not know whether the examples in GLeMM are representative of these phenomena, given their number, they provide enough data to identify trends and formulate hypotheses.

\subsection{Affix rivalry}
\label{sec:use-case-affix-rivalry}

Affix rivalry ---~which occurs when a single morphosemantic relation can be expressed by multiple affixes or derivational processes~--- is one of the most extensively studied phenomena in derivational morphology \citep{rn1679,huyghe2023.affix,salvadori2023.polyfunctional}.
In inflectional morphology, this issue is primarily addressed through inflectional classes. ``Residual'' cases fall under overabundance \citep{thornton2012.overabundance} and are rare compared to what is observed in derivation.

A well-known case of affix rivalry in English is the possibility of using \emph{\mbox{-ness}} or \emph{\mbox{-ity}} to form nouns of quality derived from adjectives \citep{marchand1969.english-wf,aronoff1976.word-formation,plag2003.word-formation-english,lindsay2013.decembrettes-2010,rn1692}.
The formation of French action nouns derived from verbs using the suffixes \emph{\mbox{-age}}, \emph{\mbox{-ment}}, and \emph{\mbox{-ion}} is part of the same phenomenon \citep{kelling2001.age-ment,fradin2019.competition}.
Similarly, the formation of relational adjectives in Russian can be achieved using the suffixes \emph{\mbox{-n-}}, \emph{\mbox{-sk-}}, and \emph{\mbox{-ov}} \citep{bobkova2023.word-structure,bobkova2025.these}
More generally, affixes that can be used to express the same morphosemantic relationship are considered rivals. For example, \emph{\mbox{-ness}} and \emph{\mbox{-ity}} form property nouns that share the same semantic relations with their base, as in \ref{exe:ness-ity-rivalry}.
\ex.\label{exe:ness-ity-rivalry}
\a.\label{exe:ness-ity-rivalry-1}
afraid\textsubscript{A} $\rightarrow$ afraidness\textsubscript{N} `the state or quality of being afraid'
\b.\label{exe:ness-ity-rivalry-2}
modern\textsubscript{A} $\rightarrow$  modernity\textsubscript{N} `the quality of being modern'

GLeMM's semantic descriptions can be used to identify some of the competing affixes.
To do this, we associate each entry in the \texttt{defs.tsv} table with a definition template obtained by replacing the lemma of the first lexeme with a placeholder <W> in the lemmatized definition of the second lexeme. 
For example, definition pattern (2d) is associated with entry \ref{exe:accuse-accusation-a}, whose alternation pattern and definition are presented in \ref{exe:accuse-accusation-b} and \ref{exe:accuse-accusation-c}.
\ex.\label{exe:accuse-accusation}
\a.\label{exe:accuse-accusation-a}	accuse, V, accusation, N
\b.\label{exe:accuse-accusation-b}	\texttt{\^{}(.+)e\$} / \texttt{\^{}(.+)ation\$}
\c.\label{exe:accuse-accusation-c}	the act of accusing
\d.\label{exe:accuse-accusation-d}	the act of <W>

Abstracting the lemma of the first lexeme provides a description of the morphosemantic relation between the two lexemes.
We then group together the pairs of lexemes that share the same part of speech (V / N in \ref{exe:accuse-accusation}) and the same definition patterns. 
The alternation patterns for these entries can be considered as rivals. For example, entries \ref{exe:accuse-accusation-a} and \ref{exe:appraise-appraisement-a} share the same definition pattern. Consequently, patterns \ref{exe:accuse-accusation-b} and  \ref{exe:appraise-appraisement-b} are potentially rivals.
\ex.\label{exe:appraise-appraisement}
\a.\label{exe:appraise-appraisement-a}	appraise, V, appraisement, N
\b.\label{exe:appraise-appraisement-b}	\texttt{\^{}(.+)\$} / \texttt{\^{}(.+)ment\$}
\c.\label{exe:appraise-appraisement-c}	the act of apprising
\d.\label{exe:appraise-appraisement-d}	the act of <W>

The rivals identified in this way include variants, since GLeMM's alternation patterns account for both allomorphs and allographs, as in \ref{exe:certify-certification}.
\ex.\label{exe:certify-certification}
\a.\label{exe:certify-certification-a}	certify, V, certification, N
\b.\label{exe:certify-certification-b}	\texttt{\^{}(.+)ify\$} /\texttt{\^{}(.+)ification\$}
\c.\label{exe:certify-certification-c}	the act of certifying
\d.\label{exe:certify-certification-d}	the act of <W>

More generally, 47 alternation patterns ---some of which are presented in (5)--- enable the realization of the definition pattern `the act of <W>'. Their exponents correspond to the main competitors of \emph{\mbox{-ation}} and \emph{\mbox{-ment}} in English. 
\ex.\label{exe:english-rivals}
\texttt{\^{}(.+)e\$} / \texttt{\^{}(.+)ation\$}\\
\texttt{\^{}(.+)\$} / \texttt{\^{}(.+)t\$}\\
\texttt{\^{}(.+)e\$} / \texttt{\^{}(.+)y\$}\\
\texttt{\^{}(.+)de\$} / \texttt{\^{}(.+)sion\$}\\
\texttt{\^{}(.+)e\$} / \texttt{\^{}(.+)tion\$}\\
\texttt{\^{}(.+)mit\$} / \texttt{\^{}(.+)mission\$}\\
\texttt{\^{}(.+)er\$} / \texttt{\^{}(.+)ry\$}\\
\texttt{\^{}(.+)\$} / \texttt{\^{}(.+)ion\$}\\
\texttt{\^{}(.+)e\$} / \texttt{\^{}(.+)al\$}\\
\texttt{\^{}(.+)t\$} / \texttt{\^{}(.+)ttal\$}\\
\texttt{\^{}(.+)vert\$} / \texttt{\^{}(.+)version\$}\\
\texttt{\^{}(.+)\$} / \texttt{\^{}(.+)al\$}\\
\texttt{\^{}(.+)e\$} / \texttt{\^{}(.+)ion\$}\\
\texttt{\^{}(.+)ify\$} / \texttt{\^{}(.+)ification\$} ...

An advantage of the consistent structure and content of the GLeMM lexicons is that an identical process can be used to collect data in its other languages.
For example, the same procedure can be used to identify the seven Italian alternation patterns in \ref{exe:italian-rivals} for the definition pattern equivalent to that in (2), namely ``azione di il <W>'' (`act of <W>').
The difference in the number of these patterns is mainly due to the comparatively small size of the GLeMM Italian lexicon.
\ex.\label{exe:italian-rivals}
\texttt{\^{}(.+)are\$} / \texttt{\^{}(.+)ezza\$}\\
\texttt{\^{}(.+)are\$} / \texttt{\^{}(.+)azione\$}\\
\texttt{\^{}(.+)re\$} / \texttt{\^{}(.+)zione\$}\\
\texttt{\^{}(.+)ire\$} / \texttt{\^{}(.+)o\$}\\
\texttt{\^{}(.+)mettere\$} / \texttt{\^{}(.+)missione\$}\\
\texttt{\^{}(.+)endere\$} / \texttt{\^{}(.+)esa\$}\\
\texttt{\^{}(.+)are\$} / \texttt{\^{}(.+)o\$}

Similarly, \ref{exe:polish-rivals} presents some of the 28 alternation patterns that, in Polish, are used to realize the definition pattern ``rzecz. odczas. od <W>'', whose expanded form is ``rzeczownik odczasownikowy od <W>'' (`noun derived from the verb <W>').
\ex.\label{exe:polish-rivals}
\texttt{\^{}(.+)ść\$} / \texttt{\^{}(.+)dnięcie\$}\\
\texttt{\^{}(.+)ić\$} / \texttt{\^{}(.+)enie\$}\\
\texttt{\^{}(.+)ścić\$} / \texttt{\^{}(.+)szczenie\$}\\
\texttt{\^{}(.+)nąć\$} / \texttt{\^{}(.+)nienie\$}\\
\texttt{\^{}(.+)wać\$} / \texttt{\^{}(.+)nie\$}\\
\texttt{\^{}(.+)ąć\$} / \texttt{\^{}(.+)ęcie\$}\\
\texttt{\^{}(.+)zić\$} / \texttt{\^{}(.+)żenie\$}\\
\texttt{\^{}(.+)eść\$} / \texttt{\^{}(.+)ecenie\$}\\
\texttt{\^{}(.+)ać\$} / \texttt{\^{}(.+)anie\$}\\
\texttt{\^{}(.+)ć\$} / \texttt{\^{}(.+)cie\$}\\
\texttt{\^{}(.+)rzeć\$} / \texttt{\^{}(.+)arcie\$}\\
\texttt{\^{}(.+)ć\$} / \texttt{\^{}(.+)nie\$}\\
\texttt{\^{}(.+)ać\$} / \texttt{\^{}(.+)enie\$}\\
\texttt{\^{}(.+)ić\$} / \texttt{\^{}(.+)jenie\$}\\
\texttt{\^{}(.+)ść\$} / \texttt{\^{}(.+)dzenie\$} ...

A similar list can be compiled from the Russian lexicon for the definition pattern ``\foreignlanguage{russian}{действие по значению гл. <W>}'', whose expanded form is ``\foreignlanguage{russian}{действие по значению глагола <W>}'' (`action corresponding to the verb <W>').
An excerpt from the 207 alternation patterns that can be used to realize this defining pattern is listed in \ref{exe:russian-rivals}.
\ex.\label{exe:russian-rivals}
\foreignlanguage{russian}{%
\texttt{\^{}(.+)ти\$} / \texttt{\^{}(.+)ение\$}\\
\texttt{\^{}(.+)ываться\$} / \texttt{\^{}(.+)\$}\\
\texttt{\^{}(.+)нуть\$} / \texttt{\^{}(.+)\$}\\
\texttt{\^{}(.+)еть\$} / \texttt{\^{}(.+)\$}\\
\texttt{\^{}(.+)аивать\$} / \texttt{\^{}(.+)ойка\$}\\
\texttt{\^{}(.+)ивать\$} / \texttt{\^{}(.+)\$}\\
\texttt{\^{}(.+)хать\$} / \texttt{\^{}(.+)шка\$}\\
\texttt{\^{}(.+)ть\$} / \texttt{\^{}(.+)тьё\$}\\
\texttt{\^{}(.+)евать\$} / \texttt{\^{}(.+)ёвка\$}\\
\texttt{\^{}(.+)иться\$} / \texttt{\^{}(.+)\$}\\
\texttt{\^{}(.+)атывать\$} / \texttt{\^{}(.+)отка\$}\\
\texttt{\^{}(.+)ивать\$} / \texttt{\^{}(.+)ение\$}\\
\texttt{\^{}(.+)епать\$} / \texttt{\^{}(.+)ёпка\$}\\
\texttt{\^{}(.+)ть\$} / \texttt{\^{}(.+)\$}\\
\texttt{\^{}(.+)гать\$} / \texttt{\^{}(.+)жение\$} ...%
}

The previous examples show how GLeMM's semantic descriptions can be used to easily compile lists of candidate competing patterns. Moreover, these candidates are globally relevant.

The definition patterns we use, clearly, do not account for the vast number of variations observed in the expression of morphosemantic relations.
Note that the goal here is not to identify all pairs of lexemes that are related in the same way, but only affixes that compete with one another among the pairs of lexemes in the GLeMM lexicons. 
A minimum number of alternation patterns that share the same definition pattern is sufficient evidence to identify their potential rivalry.
Another way to identify these rivals would have been to use vector representations of the definition patterns and to consider as rivals the alternation patterns that realize definition patterns whose representations are sufficiently similar.

\subsection{Back-formation}
\label{sec:use-case-back-formation}

Collecting examples of typical (i.e., canonical) word-formation like the \emph{\mbox{-ness}} suffixation in English, is relatively simple, provided one uses a lexicon that lists the affixes of derived lexemes.
Conversely, identifying patterns associated with atypical phenomena like back-formation is more complex because of the form-meaning discrepancy.
As we indicated in Section~\ref{sec:results-comparison-wf-relations}, back-formation refers to the derivation of a lexeme2 from a lexeme1 such that, from a formal point of view, lexeme2 could be considered a base of lexeme1.
This is the case with the example commonly used to illustrate back-formation, \emph{babysitter}\textsubscript{N} $\rightarrow$ \emph{babysit}\textsubscript{V}, which is formally indistinguishable from a regular suffixation in \emph{\mbox{-er}}, as in \emph{handcraft}\textsubscript{V}\footnote{
  Unlike \emph{babysit}\textsubscript{V}, the verb here is a conversion of the noun \emph{handcraft} and is therefore not back-formed.%
} $\rightarrow$ \emph{handcrafter}\textsubscript{N}.
Back-formation is therefore a word-formation where the direction of semantic and formal relations is reversed. 

GLeMM can be used to identify pairs of lexemes that correspond to back-formations, such as \emph{helicopter}\textsubscript{N}  $\rightarrow$ \emph{helicopt}\textsubscript{V}.
The idea is to treat as back-formations those pairs of lexemes whose orientation in \texttt{defs.tsv} is opposite to the majority orientation of the two expressions that make up their alternation patterns.
For example, the vast majority of English pairs consisting of a verb and a noun whose alternation patterns are  \texttt{\^{}(.+)\$} and \texttt{\^{}(.+)er\$}, respectively, are oriented in the direction \texttt{\^{}(.+)\$}  $\rightarrow$ \texttt{\^{}(.+)er\$} (e.g., \emph{speak}\textsubscript{V}  $\rightarrow$ \emph{speaker}\textsubscript{N}).
This order is the reverse of that of a pair like \emph{biomarker}\textsubscript{N} $\rightarrow$ \emph{biomark}\textsubscript{V} because the definition of  \emph{biomark}\textsubscript{V} is `to function as a biomarker, indicating a biological state or disease condition'. As a result, its alternation pattern  is oriented in the direction \texttt{\^{}(.+)er\$} $\rightarrow$  \texttt{\^{}(.+)\$}. Consequently,  \emph{biomarker}\textsubscript{N} $\rightarrow$ \emph{biomark}\textsubscript{V} is a back-formation.  

More generally, an $L2$ lexeme can be considered to be back-formed from an $L1$ lexeme if their alternation pattern exponent1 / exponent2 is oriented in the opposite direction of the default orientation, that is, if the number of lexeme pairs whose alternation pattern is exponent1 $\rightarrow$  exponent2 is smaller than that of lexeme pairs whose alternation pattern is exponent2 $\rightarrow$  exponent1. 
The identification could be made more precise by requiring that the lemma of L1 be longer than that of L2.
Accordingly, back-formation is detected in this procedure not as a formal anomaly, but as a reversal of a dominant derivational pattern.

This method enables the retrieval of a large number of back-formed derivatives. The French examples in \ref{exe:french-back-formation-1} follow the alternation pattern \texttt{\^{}(.+)ion\$} $\rightarrow$ \texttt{\^{}(.+)er\$}, which appears in 52 entries in the \texttt{defs.tsv} table.
This number is lower than the number of entries with the alternation pattern \texttt{\^{}(.+)er\$} $\rightarrow$ \texttt{\^{}(.+)ion\$} (65 entries).
\ex.\label{exe:french-back-formation-1}
\a.	prédation\textsubscript{N} / prédater\textsubscript{V}		`predation' / `prey on'
\b.	décoction\textsubscript{N} / décocter\textsubscript{V}	`decoction' / `decoct'
\c.	intuition\textsubscript{N} / intuiter\textsubscript{V}		`intuition' / `use intuition'

The examples in \ref{exe:french-back-formation-1} are clearly perceived as back-formed by French speakers. The same is true of the pairs in \ref{exe:french-back-formation-2}, where the alternation pattern \texttt{\^{}(.+)age\$} $\rightarrow$ \texttt{\^{}(.+)er\$}, occurring in 154 entries, is less frequent than the pattern \texttt{\^{}(.+)er\$} $\rightarrow$ \texttt{\^{}(.+)age\$}, which appears in 1,397 entries.
\ex.\label{exe:french-back-formation-2}
\a.	alpage\textsubscript{N} / alper\textsubscript{V}		`mountain pasture' / `mountain pasture'
\b.	covoiturage\textsubscript{N} / covoiturer\textsubscript{V}	`carpool' / `carpool'
\c.	engrenage\textsubscript{N} / engrener\textsubscript{V}	`gears' / `gear'

The same procedure can be used to find English examples of verbs back-formed from nouns, as in \ref{exe:english-back-formation-1} and \ref{exe:english-back-formation-2}.
The pattern in examples \ref{exe:english-back-formation-1} \texttt{\^{}(.+)ion\$} $\rightarrow$ \texttt{\^{}(.+)e\$} (465 entries) is less frequent than the pattern \texttt{\^{}(.+)e\$} $\rightarrow$ \texttt{\^{}(.+)ion\$} (581 entries). Similarly, the examples \ref{exe:english-back-formation-2} follow the pattern \texttt{\^{}(.+)er\$} $\rightarrow$ \texttt{\^{}(.+)\$} (102 entries), which is less frequent than the reverse pattern \texttt{\^{}(.+)\$} $\rightarrow$ \texttt{\^{}(.+)er\$} (2,632 entries).
\ex.\label{exe:english-back-formation-1}
\a.	allocution\textsubscript{N} / allocute\textsubscript{V}
\b.	aberration\textsubscript{N} / aberrate\textsubscript{V}

\ex.\label{exe:english-back-formation-2}
\a.	helicopter\textsubscript{N} / helicopt\textsubscript{V}
\b.	transponder\textsubscript{N} / transpond\textsubscript{V}

Another way to retrieve back-formed derivatives is to combine the entries from  GLeMM and MorphyNet, which tends to describe back-formations as regular suffixations.
As a result, a pair of lexemes that appears in opposite orientations in GLeMM and MorphyNet may correspond to a back-formation, since the orientation in MorphyNet is determined formally. 

\subsection{Symmetry}
\label{sec:use-case-symmetry}

GLeMM can be used to address the issue of symmetry in derivational morphological relations. In inflection, it is generally assumed that all forms of a lexeme have the same complexity \citep{namer2025.elsevier}.
We can therefore consider that the inflectional relations between these forms are undirected and that the inflectional graph they induce is a complete graph (each form is related to all the others).
GLeMM suggests that derivational morphological relations differ from inflectional relations on both levels. Let us first consider the orientation of the relations.

The GLeMM lexicons contain examples of lexeme pairs with a symmetrical relation. This is the case with the nouns \emph{coton}\textsubscript{N} `cotton' and \emph{cotonnier}\textsubscript{N} `cotton plant' in French, each noun having a definition that includes the other one  \ref{exe:french-symmetry}. Example \ref{exe:spanish-symmetry} presents a similar case in Spanish.

\ex.\label{exe:french-symmetry} \emph{coton}\textsubscript{N} $\rightarrow$ \emph{cotonnier}\textsubscript{N} \& \emph{cotonnier}\textsubscript{N} $\rightarrow$ \emph{coton}\textsubscript{N}
\a.	Fibre textile végétale, issue de la bourre composée de filaments longs, fins, soyeux, qui enveloppe les graines du cotonnier.\\
`A plant fiber derived from the fluff composed of long, fine, silky filaments that envelop the seeds of the cotton plant.'
\b.	Arbuste de la famille des malvacées, aux feuilles palmées, dont les fruits, des capsules ont des fibres de coton blanches autour des graines riches en huile (oléagineuses) et en protéines.\\
`A shrub belonging to the Malvaceae family, with palmate leaves, whose fruits, capsules, have white cotton fibers around seeds rich in oil (oleaginous) and protein.'

\ex.\label{exe:spanish-symmetry} \emph{abrigo}\textsubscript{N} `shelter' $\rightarrow$ \emph{abrigar}\textsubscript{V} `shelter' \& \emph{abrigar}\textsubscript{V} $\rightarrow$  \emph{abrigo}\textsubscript{N}
\a.	Acción o efecto de abrigar o de abrigarse.\\
`The act or effect of sheltering or sheltering oneself.'
\b.	Ponerse al abrigo de un cabo, costa o isla.\\
`Take shelter behind a cape, coast, or island.'

Table~\ref{tab:symmetry} presents the number of mutually motivated lexeme pairs in the GLeMM lexicons.
It shows that this configuration is rare in dictionaries (only 1\% to 4\% of pairs are symmetrical). 
These figures suggest that derivational relations actually tend to be oriented and differ from inflectional relations in this respect. They refute the hypothesis put forward by \cite{hathout2014.lilt} that derivational relations are symmetrical.
\begin{table}[th]
  \centering
  \caption{Number of mutually motivated pairs in GLeMM lexicons}
  \label{tab:symmetry}
  \begin{tabular}{lrrr}
    \hline
 & \textbf{mutually motivated pairs} & \textbf{defined pairs} & \textbf{ratio}\\
    \hline
\textbf{English} & 4,368 & 256,712 & 0.017\\
\textbf{French} & 2,422 & 77,897 & 0.031\\
\textbf{German} & 650 & 40,320 & 0.016\\
\textbf{Italian} & 314 & 9,000 & 0.035\\
\textbf{Polish} & 1,140 & 32,103 & 0.036\\
\textbf{Russian} & 3,178 & 252,014 & 0.013\\
\textbf{Spanish} & 656 & 17,483 & 0.038\\
    \hline
  \end{tabular}
\end{table}

GLeMM therefore provides a means of testing specific theoretical hypotheses against broad-ranging lexical data. While the lexicographic corpus may not be fully representative of actual usage, it does allow for the identification of clear trends, as seen here.

\subsection{Triangles}
\label{sec:use-case-triangles}

GLeMM can be used to observe another type of structure that one would expect to find if derivational paradigms defined complete graphs.
\cite{lignon2014.cmlf} proposed a paradigmatic description based on triangles, specifically triples of lexemes pairwise-connected, such as (\textrm{homogène}\textsubscript{ADJ} `homogeneous', \emph{homogénéiser}\textsubscript{V} `homogenize', \emph{homogénéisation}\textsubscript{N} `homogenization').
\cite{namer2026.compendium} suggest viewing triangles as basic building blocks that can be combined to form larger structures.
From a structural perspective, triangles take two forms (i.e., are of two types): they are either transitive relations $A \rightarrow B$ \& $B \rightarrow C$ \& $A \rightarrow C$ (15), or cycles $A \rightarrow B$ \& $B \rightarrow C$ \& $C \rightarrow A$ (15bis).

\ex.\label{exe:transitive}
\a. English\\
\emph{invent}\textsubscript{V} $\rightarrow$ \emph{invention}\textsubscript{N} $\rightarrow$ \emph{inventive}\textsubscript{ADJ} \& \emph{invent}\textsubscript{V} $\rightarrow$ \emph{inventive}\textsubscript{ADJ}
\b. French\\
\emph{voyage}\textsubscript{N} $\rightarrow$ voyager $\rightarrow$ voyageur\textsubscript{N} \& voyage\textsubscript{N} $\rightarrow$ voyageur\textsubscript{N}\\
`travel' `travel' `traveler'
 \c. German\\
\emph{dank}\textsubscript{N} $\rightarrow$ \emph{bedanken}\textsubscript{V} $\rightarrow$ \emph{unbedankt}\textsubscript{ADJ} \& \emph{dank}\textsubscript{N} $\rightarrow$ \emph{unbedankt}\textsubscript{ADJ}\\
`thanks' `thank' `not thanked'
\d. Italian\\
\emph{triste}\textsubscript{ADJ} $\rightarrow$ \emph{tristezza}\textsubscript{N} $\rightarrow$ \emph{tristemente}\textsubscript{ADV} \& \emph{triste}\textsubscript{ADJ} $\rightarrow$ \emph{tristemente}\textsubscript{ADV}\\
`sad' `sadness' `sadly'

\ex.\label{exe:cycle}
\a. English\\
\emph{skate}\textsubscript{V} $\rightarrow$ \emph{skater}\textsubscript{N} $\rightarrow$ \emph{skating}\textsubscript{N} \& \emph{skating}\textsubscript{N} $\rightarrow$ \emph{skate}\textsubscript{V}
\b. French\\
\emph{étalon}\textsubscript{N} $\rightarrow$ \emph{étalonner}\textsubscript{V} $\rightarrow$ \emph{étalonnage\textsubscript{N}} \& \emph{étalonnage}\textsubscript{N} $\rightarrow$ \emph{étalon}\textsubscript{N}\\
`standard' `calibrate' `calibration'

Tables~\ref{tab:transitive} and~\ref{tab:cycles} show the number of the two varieties of triangles in GLeMM.
They compare the number of two-edge paths  ($A \rightarrow B$, $B \rightarrow C$) with the number of triples.
The proportion of transitive relations is low, ranging from 13\% for Spanish to 2\% for German.
Circuits, on the other hand, are exceptional.
The proportion of paths of length 2 included in a circuit ranges from 1.5\% for Italian to 0.1\% for German.
\begin{table}[th]
  \centering
  \caption{Number of transitive relations in the GLeMM lexicons}
  \label{tab:transitive}
  \begin{tabular}{lrrr}
    \hline
& \textbf{transitive pairs} & \textbf{two-edge paths} & \textbf{ratio}\\
    \hline
\textbf{English} & 10,573 & 179,290 & 0.059\\
\textbf{French} & 4,557 & 40,288 & 0.113\\
\textbf{German} & 404 & 17,305 & 0.023\\
\textbf{Italian} & 175 & 1,987 & 0.088\\
\textbf{Polish} & 857 & 11,977 & 0.072\\
\textbf{Russian} & 8,508 & 206,200 & 0.041\\
\textbf{Spanish} & 752 & 5,604 & 0.134\\
    \hline
  \end{tabular}
\end{table}
\begin{table}[th]
  \centering
  \caption{Number of 3-edge cycles in GLeMM lexicons}
  \label{tab:cycles}
  \begin{tabular}{lrrr}
    \hline
& \textbf{cycles} & \textbf{two-edge paths} & \textbf{ratio}\\
    \hline
\textbf{English} & 702 & 179,290 & 0.004\\
\textbf{French} & 390 & 40,288 & 0.010\\
\textbf{German} & 24 & 17,305 & 0.001\\
\textbf{Italian} & 30 & 1,987 & 0.015\\
\textbf{Polish} & 141 & 11,977 & 0.012\\
\textbf{Russian} & 762 & 206,200 & 0.004\\
\textbf{Spanish} & 39 & 5,604 & 0.007\\
    \hline
  \end{tabular}
\end{table}

These two tables show that triangles are rare and provide further evidence to support the observations made in the previous section.
They suggest that paradigmatic families are not complete graphs. 
These findings are consistent with the proposal by \cite{hathout2025.backformation} that these families are star-shaped structures (i.e., graphs formed around a center), and with the proposal by \cite{kyjanek2020.uder} that morphological families are rooted trees.
The figures in Table~\ref{tab:transitive} also raise three questions:
Are double affixes like \emph{\mbox{-ization}} constructions on their own?
To what extent are they compositional?
How do they relate to the simple constructions that can be used to analyze them?

GLeMM therefore renews the empirical approach of morphology \citep{hathout2008.jfls,hathout2009.amf,hathout2016.cilsl}.
It opens a ``window'' enabling the observation of derivational graphs.
These observations challenge certain theoretical assumptions and pave the way for further research, even though the lack of a definition is more of a clue than proof.

\subsection{Stolons}
\label{sec:use-case-stolons}

\cite{hathout2025.backformation} proposed the concepts of stolons and pseudo-stolons to describe particular back-formations in French, such as the verb \emph{cryograver} `cryograph', formed from the noun \emph{cryogravure} `cryograph'.
A stolon is a relation between two paradigmatic families that belong to the same paradigm, as in \ref{exe:stolon-1}, where contrasts in form and content between the lexemes of the two families are identical.
\ex.\label{exe:stolon-1}
\a.	produire\textsubscript{V},	production\textsubscript{N},	producteur\textsubscript{N},	productrice\textsubscript{N},	productif\textsubscript{ADJ}\\
`produce', `production', `producer.M', `producer.F', `productive'
\b.	reproduire\textsubscript{V},	reproduction\textsubscript{N},	reproducteur\textsubscript{N},	reproductrice\textsubscript{N},	reproductif\textsubscript{ADJ}\\
`reproduce', `reproduction', `breeder.M', `breeder.F', `reproductive'

Note that while one meaning of \emph{reproduire}\textsubscript{V} is `to produce again', none of the meanings of \emph{reproducteur}\textsubscript{N} can be glossed as `one who produces again'.
The concept of a pseudo-stolon refers to a situation where two paradigmatic families are aligned formally but not semantically, as the contrasts in the content of their lexemes differ.

The concept of stolons is relatively recent. While some of the processes that give rise to them have been studied in detail \citep{dal202.eco-lave-plus-vert}, the concept of stolons \emph{per se} has not yet been the subject of a systematic study.
GLeMM provides examples that help explore these relations.
These examples are retrieved through an approximate identification of paradigmatic families, based on the assumption that the family of a lexeme L consists of all lexemes related to it either directly or via a two-edge path.
 We can then look for subsets of these families of a given size with identical alternation patterns. Examples \ref{exe:spanish-stolon} in Spanish and \ref{exe:french-stolon} in French illustrate these structures.
\ex.\label{exe:spanish-stolon}
 \begin{tabular}[t]{ll}
cargar\textsubscript{V} `load' & descargar\textsubscript{V} `unload' \\
carga\textsubscript{N}`load.OBJ' & descarga\textsubscript{N} `unload.OBJ'\\
cargador\textsubscript{N} `loader' & descargador\textsubscript{N} `unloader'\\
cargarse\textsubscript{V} `load.PRONOMINAL' & descargarse\textsubscript{V} `unload.PRONOMINAL'
 \end{tabular}

\ex.\label{exe:french-stolon}
 \begin{tabular}[t]{ll}
charger\textsubscript{V} `load' & décharger\textsubscript{V} `unload' \\
autocharger\textsubscript{V} `autoload' & autodécharger\textsubscript{V} `autounload'\\
charge\textsubscript{N} `load.OBJ' & décharge\textsubscript{N} `unload.OBJ' \\
chargeable\textsubscript{ADJ} `loadable' & déchargeable\textsubscript{ADJ} `unloadable'\\
chargeage\textsubscript{N} `loading' & déchargeage\textsubscript{N} `unloading'\\
chargement\textsubscript{N} `loading' & déchargement\textsubscript{N} `unloading' \\
chargeur\textsubscript{N} `loader' & déchargeur\textsubscript{N} `unloader' \\
chargé\textsubscript{ADJ} `loaded' & déchargé\textsubscript{ADJ} `unloaded' \\
recharger\textsubscript{V} `reload' & redécharger\textsubscript{V} `reunload'
 \end{tabular}

These examples point to the fact that, from a formal perspective, stolons primarily correspond to prefixation. They highlight the asymmetry between suffixation and prefixation.
More generally, stolons seem to be induced by a relatively small number of prefixes (such as \emph{\mbox{re-}} and \emph{\mbox{dé-}} in French) with a modal meaning (e.g., `again' for \emph{\mbox{re-}}) or an adverbial meaning (e.g., `oneself' for \emph{\mbox{auto-}}).

\section{Conclusion}
\label{sec:conclusion}

In this article, we have introduced GLeMM, a new resource that contains derivational lexicons for seven European languages,  designed for experimentation and corpus-based description.
The potential of this resource  and the wide range of issues it can address has been illustrated by the use cases in Section~\ref{sec:use-cases}. 
What makes this resource fundamentally different are the semantic descriptions provided in the \texttt{defs.tsv} tables, which, as we have seen, offer a ``window'' into the paradigmatic organization of derivational morphology.
To the best of our knowledge, no other resource offers comparable descriptions.
Although they do not follow any specific formal structure, they enable observations and experiments that are beyond the scope of other resources.

GLeMM also stands out for the size of its lexicons, which exceed the size of many other resources, and for the diversity of linguistic phenomena it covers, suggesting that its lexicons may be fairly representative. 
This results from the fact that the lexicographic descriptions used to create it were not intended for morphological analysis and instead reflect general lexical relationships and patterns. 
Furthermore, the retrieval of derivational relations in dictionaries is systematic, which minimizes the potential biases that may exist in manually created resources.
Another advantage of this systematic approach is the consistency of these descriptions.

GLeMM has significant room for improvement in terms of the number of languages it covers and of the richness of its annotations. New editions of Wiktionary are regularly added to the Kaikki project, enabling the creation of GLeMM lexicons for those languages.  We plan to add lexicons for Czech, Dutch, Greek, Portuguese, and Turkish in the near future.

Another direction in which we plan to improve is the description of morphophonology \citep{lignon2025.phononette}.  This will involve (\emph{i}) using phonemic transcriptions as formal descriptions and (\emph{ii}) applying FAPinette to inflectional morphology to obtain alternation patterns similar to those we use for derivation.

The FAPinette method used to create GLeMM lexicons makes no distinction between the different types of word-formation. Derivation and compounding are dealt with in the same way.  Whenever a compounding element is used to form a large number of compounds, an alternation pattern covering them is abstracted. This approach is not entirely satisfactory for several reasons
\begin{itemize}
\item Compounds belong to multiple word families. The connections they create between these families make them harder to separate.
\item The large number of compounding elements distinguishes compounding from derivation, which involves a smaller number of affixes. This factor has a direct impact on the FAPinette settings.
\item The structures formed by compounds and derivatives are distinct yet interconnected. GLeMM's lexicons describe them in a single graph mixing the two.
\item The prevalence of compounds in the lexicons varies across languages (e.g., they are less frequent in French than in German). These differences may bias the observation of particular derivational phenomena.
\end{itemize}
We plan to annotate GLeMM entries by specifying the types of processes.  This will require an analysis of the etymological sections of Wiktionary entries.

As outlined in Section~\ref{sec:results-comparison-wf-relations}, GLeMM entries cover a wide variety of morphological phenomena.
However, the structure of the lexicons (i.e., graphs of lexemes) and the annotations of their entries cannot easily be used to identify some of these phenomena, particularly because the meaning of the lexemes is not described using a formal language. 
For example, a parasynthetic construction is formally indistinguishable from double affixation.
To overcome this limitation, entries must be annotated using better-controlled semantic representations, such as standardized definitions.
Another avenue that remains to be explored is that of a properly paradigmatic description of the GLeMM lexicons.


\begin{thebibliography}{117}
\providecommand{\natexlab}[1]{#1}
\providecommand{\url}[1]{#1}
\providecommand{\urlprefix}{}
\expandafter\ifx\csname urlstyle\endcsname\relax
  \providecommand{\doi}[1]{doi:\discretionary{}{}{}#1}\else
  \providecommand{\doi}{doi:\discretionary{}{}{}\begingroup
  \urlstyle{rm}\Url}\fi

\bibitem[{Ackerman et~al.(2009)Ackerman, Blevins \&
  Malouf}]{ackerman2009.analogy}
Ackerman, Farrell, James~P Blevins \& Robert Malouf. 2009.
\newblock Parts and wholes: Implicative patterns in inflectional paradigms.
\newblock In James~P Blevins \& Juliette Blevins (eds.), \emph{Analogy in
  grammar: Form and acquisition}, 54--81. Oxford: Oxford University Press.

\bibitem[{Albright \& Hayes(2003)}]{albright2003.mgl}
Albright, Adam \& Bruce Hayes. 2003.
\newblock Rules vs. analogy in {E}nglish past tenses: a
  computational/experimental study.
\newblock \emph{Cognition} 90(2). 119--161.
\newblock \doi{https://doi.org/10.1016/S0010-0277(03)00146-X}.

\bibitem[{Arndt-Lappe(2014)}]{rn1692}
Arndt-Lappe, Sabine. 2014.
\newblock Analogy in suffix rivalry: the case of english \emph{\mbox{-ity}} and
  \emph{\mbox{-ness}}.
\newblock \emph{English Language and Linguistics} 18. 497--548.

\bibitem[{Aronoff(1976)}]{aronoff1976.word-formation}
Aronoff, Mark. 1976.
\newblock \emph{Word formation in generative grammar} Linguistic Inquiry
  Monographs.
\newblock Cambridge, MA: MIT Press.

\bibitem[{Aronoff(2019)}]{rn1679}
Aronoff, Mark. 2019.
\newblock Competitors and alternants in linguistic morphology.
\newblock In Franz Rainer, Wolfgang~U. Dressler \& Hans~Christian Luschützky
  (eds.), \emph{Competition in inflection and word-formation}, 39--66.
  Springer.

\bibitem[{Baayen et~al.(1995)Baayen, Piepenbrock \&
  Gulikers}]{baayen1995.celex}
Baayen, R.~Harald, Richard Piepenbrock \& Leon Gulikers. 1995.
\newblock The {CELEX} lexical database (release 2).
\newblock CD-ROM.
\newblock Linguistic Data Consortium, Philadelphia, PA.

\bibitem[{Bagasheva(2017)}]{bagasheva2017.semantic-concepts}
Bagasheva, Alexandra. 2017.
\newblock Comparative semantic concepts in affixation.
\newblock In Santana~Lario Juan \& Salvador Valera (eds.), \emph{Competing
  patterns in {E}nglish affixation}, 33--65. Peter Lang Bern.

\bibitem[{Barque et~al.(2020)Barque, Haas, Huyghe, Tribout, Candito, Crabbé \&
  Segonne}]{rn1661}
Barque, Lucie, Pauline Haas, Richard Huyghe, Delphine Tribout, Marie Candito,
  Benoit Crabbé \& Vincent Segonne. 2020.
\newblock {FrSemCor}: Annotating a {F}rench corpus with supersenses.
\newblock In \emph{12th edition of its language resources and evaluation
  conference ({LREC})}, ELRA.
\newblock \urlprefix\url{https://hal.archives-ouvertes.fr/hal-02511929}.

\bibitem[{Batsuren et~al.(2019)Batsuren, Bella \&
  Giunchiglia}]{batsuren2019.cognet}
Batsuren, Khuyagbaatar, Gabor Bella \& Fausto Giunchiglia. 2019.
\newblock {C}og{N}et: A large-scale cognate database.
\newblock In \emph{Proceedings of the 57th annual meeting of the association
  for computational linguistics}, 3136--3145. Florence, Italy.

\bibitem[{Batsuren et~al.(2021)Batsuren, Bella \&
  Giunchiglia}]{batsuren2021.morphynet}
Batsuren, Khuyagbaatar, G{\'a}bor Bella \& Fausto Giunchiglia. 2021.
\newblock {M}orphy{N}et: a large multilingual database of derivational and
  inflectional morphology.
\newblock In \emph{Proceedings of the 18th {SIGMORPHON} workshop on
  computational research in phonetics, phonology, and morphology}, 39--48.

\bibitem[{Batsuren et~al.(2022)Batsuren, Goldman \&
  al}]{batsuren2022.unimorph-short}
Batsuren, Khuyagbaatar, Omer Goldman \& al. 2022.
\newblock {U}ni{M}orph 4.0: {U}niversal {M}orphology.
\newblock In \emph{Proceedings of the thirteenth language resources and
  evaluation conference}, 840--855. Marseille, France.

\bibitem[{Bauer(2017)}]{bauer2017.compounding}
Bauer, Laurie. 2017.
\newblock \emph{Compounds and compounding}.
\newblock Cambridge University Press.

\bibitem[{Beniamine(2018)}]{beniamine2018.phd}
Beniamine, Sacha. 2018.
\newblock \emph{Classifications flexionnelles. {{\'E}}tude quantitative des
  structures de paradigmes}: Univeristé Paris Diderot Thèse de doctorat.

\bibitem[{Beniamine \& Naranjo(2021)}]{beniamine2021.SCIL}
Beniamine, Sacha \& Mat{\'i}as~Guzm{\'a}n Naranjo. 2021.
\newblock Multiple alignments of inflectional paradigms.
\newblock In \emph{Proceedings of the society for computation in linguistics
  2021}, 216--227.

\bibitem[{Bobkova(2025)}]{bobkova2025.these}
Bobkova, Natalia. 2025.
\newblock \emph{La concurrence suffixale dans la construction des adjectifs
  dénominaux en russe : analyse des suffixes \emph{\mbox{-n-}},
  \emph{\mbox{-sk-}} et \emph{\mbox{-ov}}}: Université de Toulouse Thèse de
  doctorat.

\bibitem[{Bobkova \& Montermini(2023)}]{bobkova2023.word-structure}
Bobkova, Natalia \& Fabio Montermini. 2023.
\newblock A quantitative approach to doublets in {R}ussian denominal adjective
  construction.
\newblock \emph{Word Structure} 16(1). 63--86.
\newblock \doi{10.3366/word.2023.0221}.

\bibitem[{Bonami \& Beniamine(2016)}]{bonami2016.ws}
Bonami, Olivier \& Sacha Beniamine. 2016.
\newblock Joint predictiveness in inflectional paradigms.
\newblock \emph{Word Structure} 9(2). 156--182.

\bibitem[{Calderone et~al.(2016)Calderone, Sajous \&
  Hathout}]{calderone2016.sle}
Calderone, Basilio, Franck Sajous \& Nabil Hathout. 2016.
\newblock {GLAW-IT}: A free large {I}talian dictionary encoded in a
  fine-grained {XML} format.
\newblock In \emph{Proceedings of the 49th annual meeting of the societas
  linguistica europaea (sle 2016)}, 43--45. Naples, Italy.

\bibitem[{Cardillo et~al.(2018)Cardillo, Ferro, Marzi \&
  Pirrelli}]{cardillo2018.pcfp}
Cardillo, Alberto~Franco, Marcello Ferro, Claudia Marzi \& Vito Pirrelli. 2018.
\newblock Deep learning of inflection and the cell-filling problem.
\newblock \emph{Italian Journal of Computational Linguistics} 4(1). 57--75.

\bibitem[{Cotterell \& Schütze(2018)}]{cotterell2018.joint}
Cotterell, Ryan \& Hinrich Schütze. 2018.
\newblock Joint semantic synthesis and morphological analysis of the derived
  word.
\newblock \emph{Transactions of the Association for Computational Linguistics}
  6. 33--48.

\bibitem[{Creutz \& Lagus(2002)}]{creutz2002.acl}
Creutz, Mathias \& Krista Lagus. 2002.
\newblock Unsupervised discovery of morphemes.
\newblock In \emph{Proceedings of the {ACL} workshop on morphological and
  phonological learning}, 21--30. Philadelphia, PA: ACL.

\bibitem[{Creutz \& Lagus(2004)}]{creutz2004.sigorphon}
Creutz, Mathias \& Krista Lagus. 2004.
\newblock Induction of a simple morphology for highly-inflecting languages.
\newblock In \emph{Proceedings of the 7th meeting of the {ACL} special interest
  group in computational phonology: Current themes in computational phonology
  and morphology}, 43--51. Barcelona, Spain.

\bibitem[{Creutz \& Lagus(2005)}]{creutz2005.morfessor}
Creutz, Mathias \& Krista Lagus. 2005.
\newblock Unsupervised morpheme segmentation and morphology induction from text
  corpora using {M}orfessor 1.0.
\newblock Tech. Rep. A81 Helsinki University of Technology.

\bibitem[{Dal \& Namer(2022)}]{dal202.eco-lave-plus-vert}
Dal, Georgette \& Fiammetta Namer. 2022.
\newblock {\emph{\'Eco-}} lave plus vert, et il lave toute la famille.
\newblock \emph{Neologica} 16. 111--128.
\newblock \doi{10.48611/isbn.978-2-406-13219-6.p.0111}.

\bibitem[{Dendien \& Pierrel(2003)}]{dendien2003.tal}
Dendien, Jacques \& Jean-Marie Pierrel. 2003.
\newblock Le {T}résor de la {L}angue {F}rançaise informatisé: un exemple
  d'informatisation d'un dictionnaire de langue de référence.
\newblock \emph{Traitement automatique des langues} 44(2). 11--37.

\bibitem[{Fellbaum(1998)}]{fellbaum98}
Fellbaum, Christiane. 1998.
\newblock \emph{Wordnet: An electronic lexical database}.
\newblock MIT Press.

\bibitem[{Fellbaum(1999)}]{fellbaum1999.wordnet}
Fellbaum, Christiane (ed.). 1999.
\newblock \emph{Wordnet: an electronic lexical database}.
\newblock Cambridge, MA: MIT Press.

\bibitem[{Fellbaum et~al.(2009)Fellbaum, Osherson \&
  Clark}]{fellbaum2009.morphosemantic-wn3}
Fellbaum, Christiane, Anne Osherson \& Peter~E. Clark. 2009.
\newblock Putting semantics into {W}ord{N}et's ``morphosemantic'' links.
\newblock In \emph{Human language technology. challenges of the information
  society}, vol. 5603 Lecture Notes in Computer Science Volume, 350--358.
  Springer.

\bibitem[{Fradin(2019)}]{fradin2019.competition}
Fradin, Bernard. 2019.
\newblock Competition in derivation: What can we learn from {F}rench doublets
  in \emph{\mbox{-age}} and \emph{\mbox{-ment}}?
\newblock In Franz Rainer, Francesco Gardani, Wolfgang~U. Dressler \&
  Hans~Christian Luschützky (eds.), \emph{Competition in inflection and
  word-formation}, 67--93. Springer.

\bibitem[{Gage(1994)}]{gage1994.BPE}
Gage, Philip. 1994.
\newblock A new algorithm for data compression.
\newblock \emph{C Users Journal} 12(2). 23–38.

\bibitem[{Goldsmith(2001)}]{goldsmith2001.cl}
Goldsmith, John. 2001.
\newblock Unsupervised learning of the morphology of natural language.
\newblock \emph{Computational Linguistics} 27(2). 153--198.

\bibitem[{Goldsmith(2006)}]{goldsmith2006.nle}
Goldsmith, John. 2006.
\newblock An algorithm for the unsupervised learning of morphology.
\newblock \emph{Natural Language Engineering} 12(4). 353--371.

\bibitem[{Guzm{\'a}n~Naranjo(2020)}]{guzman2020.analogy}
Guzm{\'a}n~Naranjo, Mat{\'\i}as. 2020.
\newblock Analogy, complexity and predictability in the {R}ussian nominal
  inflection system.
\newblock \emph{Morphology} 30(3). 219--262.

\bibitem[{Habash \& Dorr(2003)}]{habash2003.catvar}
Habash, Nizar \& Bonnie Dorr. 2003.
\newblock A categorial variation database for {E}nglish.
\newblock In \emph{Proceedings of the human language technology and north
  american association for computational linguistics conference (naacl/hlt
  2003)}, 96--102. Edmonton: ACL.

\bibitem[{Hathout(2001)}]{hathout2001.taln}
Hathout, Nabil. 2001.
\newblock Analogies morpho-synonymiques. {U}ne méthode d'acquisition
  automatique de liens morphologiques à partir d'un dictionnaire de synonymes.
\newblock In Denis Maurel (ed.), \emph{Actes de la 8\ieme\ conférence annuelle
  sur le traitement automatique des langues naturelles (taln-2001)}, 223--232.
  Tours: ATALA.

\bibitem[{Hathout(2002)}]{hathout2002.lrec.wordnet}
Hathout, Nabil. 2002.
\newblock From {WordNet} to {CELEX}: {A}cquiring morphological links from
  dictionaries of synonyms.
\newblock In \emph{Proceedings of the third international conference on
  language resources and evaluation}, 1478--1484. Las Palmas de Gran Canaria:
  ELRA.

\bibitem[{Hathout(2005)}]{hathout2005.cahlex}
Hathout, Nabil. 2005.
\newblock Exploiter la structure analogique du lexique construit : {U}ne
  approche computationnelle.
\newblock \emph{Cahiers de lexicologie} 87(2). 5--28.

\bibitem[{Hathout(2008)}]{hathout2008.textgraphs3}
Hathout, Nabil. 2008.
\newblock Acquisition of the morphological structure of the lexicon based on
  lexical similarity and formal analogy.
\newblock In \emph{Proceedings of the coling workshop textgraphs-3}, 1--8.
  Manchester: {ACL}.

\bibitem[{Hathout(2009{\natexlab{a}})}]{hathout2009.taln}
Hathout, Nabil. 2009{\natexlab{a}}.
\newblock Acquisition morphologique à partir d'un dictionnaire informatisé.
\newblock In \emph{Actes de la 16\ieme\ conférence sur le traitement
  automatique des langues naturelles (taln-2009)}, Senlis: ATALA.

\bibitem[{Hathout(2009{\natexlab{b}})}]{hathout2009.decembrettes}
Hathout, Nabil. 2009{\natexlab{b}}.
\newblock Acquisition of morphological families and derivational series from a
  machine readable dictionary.
\newblock In Fabio Montermini, Gilles Boyé \& Jesse Tseng (eds.),
  \emph{Selected proceedings of the 6th décembrettes: Morphology in bordeaux},
  Somerville, MA: Cascadilla Proceedings Project.

\bibitem[{Hathout(2011{\natexlab{a}})}]{hathout2011.lel}
Hathout, Nabil. 2011{\natexlab{a}}.
\newblock Morphonette: a paradigm-based morphological network.
\newblock \emph{Lingue e linguaggio} 2011(2). 243--262.

\bibitem[{Hathout(2011{\natexlab{b}})}]{hathout2011.dumal}
Hathout, Nabil. 2011{\natexlab{b}}.
\newblock Une approche topologique de la construction des mots : propositions
  théoriques et application à la préfixation en \mbox{\emph{anti-}}.
\newblock In Michel Roché, Gilles Boyé, Nabil Hathout, Stéphanie Lignon \&
  Marc Plénat (eds.), \emph{Des unités morphologiques au lexique}, 251--318.
  Hermès Science-Lavoisier.

\bibitem[{Hathout(2014)}]{hathout2014.ls}
Hathout, Nabil. 2014.
\newblock Phonotactics in morphological similarity metrics.
\newblock \emph{Language Sciences} 46. 71--83.

\bibitem[{Hathout(2016)}]{hathout2016.cilsl}
Hathout, Nabil. 2016.
\newblock La question des données en morphologie.
\newblock \emph{Cahiers de l'ILSL} 45. 123--160.

\bibitem[{Hathout et~al.(2025)Hathout, Calderone, Sajous \&
  Namer}]{hathout2025.form-meaning}
Hathout, Nabil, Basilio Calderone, Franck Sajous \& Fiammetta Namer. 2025.
\newblock Form and meaning in word-formation: Who does what?
\newblock Manuscript.

\bibitem[{Hathout et~al.(2008)Hathout, Montermini \& Tanguy}]{hathout2008.jfls}
Hathout, Nabil, Fabio Montermini \& Ludovic Tanguy. 2008.
\newblock Extensive data for morphology: {U}sing the {W}orld {W}ide {W}eb.
\newblock \emph{Journal of {F}rench Language Studies} 18(1). 67--85.

\bibitem[{Hathout \& Namer(2014)}]{hathout2014.lilt}
Hathout, Nabil \& Fiammetta Namer. 2014.
\newblock Démonette, a {F}rench derivational morpho-semantic network.
\newblock \emph{Linguistic Issues in Language Technology} 11(5). 125--168.

\bibitem[{Hathout \& Namer(2016)}]{hathout2016.lrec-demonette}
Hathout, Nabil \& Fiammetta Namer. 2016.
\newblock Giving lexical resources a second life: {D}émonette, a multi-sourced
  morpho-semantic network for {F}rench.
\newblock In \emph{Proceedings of the tenth international conference on
  language resources and evaluation ({{LREC}} 2016)}, Portorož, Slovenia.

\bibitem[{Hathout \& Namer(2018)}]{hathout2018.hommage-fradin}
Hathout, Nabil \& Fiammetta Namer. 2018.
\newblock La parasynthèse à travers les modèles~: des {RCL} au {P}ara{D}is.
\newblock In Olivier Bonami, Gilles Boyé, Georgette Dal, Hélène Giraudo \&
  Fiammetta Namer (eds.), \emph{The lexeme in descriptive and theorical
  morphology}, 365--399. Langage Sciences Press.

\bibitem[{Hathout \& Namer(2025)}]{hathout2025.backformation}
Hathout, Nabil \& Fiammetta Namer. 2025.
\newblock What do derivational paradigms tell us about back-formation and what
  does back-formation tell us about derivational paradigms?
\newblock \emph{Word Structure} 18(3). 239--280.

\bibitem[{Hathout et~al.(2009)Hathout, Namer, Plénat \&
  Tanguy}]{hathout2009.amf}
Hathout, Nabil, Fiammetta Namer, Marc Plénat \& Ludovic Tanguy. 2009.
\newblock La collecte et l'utilisation des données en morphologie.
\newblock In Bernard Fradin, Françoise Kerleroux \& Marc Plénat (eds.),
  \emph{Aperçus de morphologie du français}, 267--287. Saint-Denis: Presses
  universitaires de Vincennes.

\bibitem[{Hathout \& Sajous(2016)}]{hathout2016.lrec-glawi}
Hathout, Nabil \& Franck Sajous. 2016.
\newblock Wiktionnaire's {W}ikicode {GLAWI}fied: a workable {F}rench
  machine-readable dictionary.
\newblock In \emph{Proceedings of the tenth international conference on
  language resources and evaluation ({{LREC}} 2016)}, Portorož, Slovenia.

\bibitem[{Hathout et~al.(2014)Hathout, Sajous \& Calderone}]{hathout2014.lg-lp}
Hathout, Nabil, Franck Sajous \& Basilio Calderone. 2014.
\newblock Acquisition and enrichment of morphological and morphosemantic
  knowledge from the {F}rench {W}iktionary.
\newblock In \emph{Proceedings of the {COLING} workshop on lexical and
  grammatical resources for language processing}, 65--74. Dublin, Ireland.

\bibitem[{Hathout et~al.(2020)Hathout, Sajous, Calderone \&
  Namer}]{hathout2020.lrec-glawinette}
Hathout, Nabil, Franck Sajous, Basilio Calderone \& Fiammetta Namer. 2020.
\newblock {G}lawinette: a linguistically motivated derivational description of
  {F}rench acquired from {GLAWI}.
\newblock In \emph{Proceedings of the twelfth international conference on
  language resources and evaluation ({{LREC}} 2020)}, 3870--3878. Marseille.

\bibitem[{Hay \& Baayen(2003)}]{hay2003.phonotactics}
Hay, Jennifer \& Harald Baayen. 2003.
\newblock Phonotactics, parsing and productivity.
\newblock \emph{Italian Journal of Linguistics} 15(1). 99–130.

\bibitem[{Hledíková \& Ševčíková(2024)}]{hledikova2024.conversion}
Hledíková, Hana \& Magda Ševčíková. 2024.
\newblock Conversion in languages with different morphological structures: a
  semantic comparison of {E}nglish and {C}zech.
\newblock \emph{Morphology} 34(1). 73--102.
\newblock \doi{10.1007/s11525-024-09422-1}.

\bibitem[{Huguin et~al.(2023)Huguin, Barque, Haas \& Tribout}]{rn1870}
Huguin, Mathilde, Lucie Barque, Pauline Haas \& Delphine Tribout. 2023.
\newblock Typage sémantique des noms dans la ressource morphologique
  {D}émonette.
\newblock \emph{Lexique} 33. 41--56.
\newblock \doi{10.54563/lexique.1086}.
\newblock \urlprefix\url{http://www.peren-revues.fr/lexique/1086}.

\bibitem[{Huyghe \& Varvara(2023)}]{huyghe2023.affix}
Huyghe, Richard \& Rossella Varvara. 2023.
\newblock Affix rivalry: Theoretical and methodological challenges.
\newblock \emph{Word Structure} 16(1). 1--23.

\bibitem[{de~Jong et~al.(2000)de~Jong, Schreuder \&
  Baayen}]{dejong2000.family-size}
de~Jong, Nivja~H., Robert Schreuder \& R.~Harald Baayen. 2000.
\newblock The morphological family size effect and morphology.
\newblock \emph{Language and cognitive processes} 15(4/5). 329--365.

\bibitem[{Kann \& Sch{\"u}tze(2016)}]{kann-schutze-2016-single}
Kann, Katharina \& Hinrich Sch{\"u}tze. 2016.
\newblock Single-model encoder-decoder with explicit morphological
  representation for reinflection.
\newblock In Katrin Erk \& Noah~A. Smith (eds.), \emph{Proceedings of the 54th
  annual meeting of the association for computational linguistics (volume 2:
  Short papers)}, 555--560. Berlin, Germany: Association for Computational
  Linguistics.

\bibitem[{Kelling(2001)}]{kelling2001.age-ment}
Kelling, Carmen. 2001.
\newblock Agentivity and suffix selection.
\newblock In \emph{Proceedings of the {LFG} conference}, 147--162. Stanford,
  CA: CSLI.

\bibitem[{Koehl(2012)}]{koehl2012.phd}
Koehl, Aurore. 2012.
\newblock \emph{La construction morphologique des noms désadjectivaux
  suffixés en français}.
\newblock Nancy: Université de {L}orraine Thèse de doctorat.

\bibitem[{Koehl \& Lignon(2014)}]{koehl2014.morphology}
Koehl, Aurore \& Stéphanie Lignon. 2014.
\newblock Property nouns with \emph{-ité} and \emph{-itude}: formal
  alternation and morphopragmatics or the sad-itude of the {A}ité$_{N}$.
\newblock \emph{Morphology} 24(4). 351--376.

\bibitem[{Kyj{\'{a}}nek(2018)}]{kyjanek2018.resources}
Kyj{\'{a}}nek, Luk{\'{a}}{\v{s}}. 2018.
\newblock Morphological resources of derivational word-formation relations.
\newblock Tech. Rep.~61 {\'{U}}FAL - Charles University Prague.

\bibitem[{Kyj{\'a}nek et~al.(2022)Kyj{\'a}nek, Lyashevskaya, Nedoluzhko,
  Vodolazsky \& {\v{Z}}abokrtsk{\'y}}]{kyjanek2022.derinet-RU}
Kyj{\'a}nek, Luk{\'a}{\v{s}}, Olga Lyashevskaya, Anna Nedoluzhko, Daniil
  Vodolazsky \& Zden{\v{e}}k {\v{Z}}abokrtsk{\'y}. 2022.
\newblock Constructing a lexical resource of {R}ussian derivational morphology.
\newblock In \emph{Proceedings of the thirteenth language resources and
  evaluation conference}, 2788--2797. Marseille, France.

\bibitem[{Kyj{\'a}nek et~al.(2021)Kyj{\'a}nek, {\v Z}abokrtsk{\'y}, Vidra \&
  {\v S}ev{\v c}{\'{\i}}kov{\'a}}]{uder-data}
Kyj{\'a}nek, Luk{\'a}{\v s}, Zden{\v e}k {\v Z}abokrtsk{\'y}, Jon{\'a}{\v s}
  Vidra \& Magda {\v S}ev{\v c}{\'{\i}}kov{\'a}. 2021.
\newblock Universal derivations v1.1.
\newblock {LINDAT}/{CLARIAH}-{CZ} digital library at the Institute of Formal
  and Applied Linguistics ({{\'U}FAL}), Faculty of Mathematics and Physics,
  Charles University.

\bibitem[{Kyjánek et~al.(2020)Kyjánek, Žabokrtský, Ševčíková \&
  Vidra}]{kyjanek2020.uder}
Kyjánek, Lukáš, Zdenĕk Žabokrtský, Magda Ševčíková \& Jonáš Vidra.
  2020.
\newblock {U}niversal {D}erivations 1.0, a growing collection of harmonised
  word-formation resources.
\newblock \emph{The Prague Bulletin of Mathematical Linguistics} 115. 5--30.

\bibitem[{Langlais \& Yvon(2008)}]{langlais2008.coling}
Langlais, Philippe \& François Yvon. 2008.
\newblock Scaling up analogical learning.
\newblock In \emph{Proceedings of the 22nd international conference on
  computational linguistics (coling 2008)}, 51–54. Manchester.

\bibitem[{Lango et~al.(2018)Lango, {\v{S}}ev{\v{c}}{\'i}kov{\'a} \&
  {\v{Z}}abokrtsk{\'y}}]{lango2018.derinet-ES-PL}
Lango, Mateusz, Magda {\v{S}}ev{\v{c}}{\'i}kov{\'a} \& Zden{\v{e}}k
  {\v{Z}}abokrtsk{\'y}. 2018.
\newblock Semi-automatic construction of word-formation networks (for {P}olish
  and {S}panish).
\newblock In \emph{Proceedings of the eleventh international conference on
  language resources and evaluation ({LREC} 2018)}, Miyazaki, Japan.

\bibitem[{Lango et~al.(2021)Lango, Žabokrtský \& Ševčíková}]{rn1925}
Lango, Mateusz, Zdenĕk Žabokrtský \& Magda Ševčíková. 2021.
\newblock Semi-automatic construction of word-formation networks.
\newblock \emph{Language Resources and Evaluation} 55. 3--32.
\newblock \doi{10.1007/s10579-019-09484-2}.

\bibitem[{Lavallée \& Langlais(2009)}]{lavallee2009.morphochallenge}
Lavallée, Jean-François \& Philippe Langlais. 2009.
\newblock Morphological acquisition by formal analogy.
\newblock In \emph{Working notes for the morphochallenge at clef 2009}, Corfu,
  Greece.

\bibitem[{Lepage(1998)}]{lepage1998.coling}
Lepage, Yves. 1998.
\newblock Solving analogies on words: {A}n algorithm.
\newblock In \emph{Proceedings of the 36th annual meeting of the association
  for computational linguistics and of the 17th international conference on
  computational linguistics}, vol.~2, 728--735. Montréal.

\bibitem[{Lepage(2003)}]{lepage2003.hdr}
Lepage, Yves. 2003.
\newblock \emph{De l'analogie rendant compte de la commutation en
  linguistique}.
\newblock Grenoble: Université Joseph Fourier Habilitation à diriger des
  recherches.

\bibitem[{Lepage(2004)}]{lepage2004.analogy-formal-languages}
Lepage, Yves. 2004.
\newblock Analogy and formal languages.
\newblock \emph{Electronic Notes in Theoretical Computer Science} 53. 180--191.
\newblock Proceedings of the the 6th Conference on Formal Grammar and the 7th
  on the Mathematics of Language (FG/MOL-2001).

\bibitem[{Lignon et~al.(2025)Lignon, Dal, Hathout \&
  Namer}]{lignon2025.phononette}
Lignon, Stéphanie, Georgette Dal, Nabil Hathout \& Fiammetta Namer. 2025.
\newblock La morphophonologie est-elle paradigmatique ? {P}hononette vous
  répond.
\newblock \emph{Langue Française} 228. 59--16.

\bibitem[{Lignon et~al.(2014)Lignon, Namer \& Villoing}]{lignon2014.cmlf}
Lignon, Stéphanie, Fiammetta Namer \& Florence Villoing. 2014.
\newblock De l'agglutination à la triangulation ou comment expliquer certaines
  séries morphologiques.
\newblock In \emph{Actes du 4\ieme\ congrès mondial de linguistique française
  ({CMLF} 2014)}, 1813--1836.

\bibitem[{Lignon \& Roché(2011)}]{lignon2011.dumal}
Lignon, Stéphanie \& Michel Roché. 2011.
\newblock Entre histoire et morphophonologie, quelle distribution pour
  \mbox{\emph{-éen}} vs \mbox{\emph{-ien}} ?
\newblock In Michel Roché, Gilles Boyé, Nabil Hathout, Stéphanie Lignon \&
  Marc Plénat (eds.), \emph{Des unités morphologiques au lexique}, 191--250.
  Hermès Science-Lavoisier.

\bibitem[{Lindsay \& Aronoff(2013)}]{lindsay2013.decembrettes-2010}
Lindsay, Mark \& Mark Aronoff. 2013.
\newblock Natural selection in self-organizing morphological systems.
\newblock In Nabil Hathout, Fabio Montermini \& Jesse Tseng (eds.),
  \emph{{M}orphology in {T}oulouse}, 133--153. München: Lincom Europa.

\bibitem[{Malouf(2017)}]{malouf2017.morphology}
Malouf, Rob. 2017.
\newblock Abstractive morphological learning with a recurrent neural network.
\newblock \emph{Morphology} 27(4). 431–458.

\bibitem[{Marchand(1969)}]{marchand1969.english-wf}
Marchand, Hans. 1969.
\newblock \emph{The categories and types of present-day {E}nglish
  word-formation: {A} synchronic-diachronic approach}.
\newblock Beck.

\bibitem[{Martin(1992)}]{martin1992.robert.pls}
Martin, Robert. 1992.
\newblock \emph{Pour une logique du sens} Linguistique nouvelle.
\newblock Paris: Presses universitaires de France.

\bibitem[{Miller et~al.(1990)Miller, Beckwith, Fellbaum, Gross \&
  Miller}]{miller1990.wordnet}
Miller, Georges~A., Richard Beckwith, Christiane Fellbaum, Derek Gross \&
  Katherine~J. Miller. 1990.
\newblock Introduction to {W}ord{N}et: An on-line lexical database.
\newblock \emph{International Journal of Lexicography} 3(4). 335--391.

\bibitem[{Namer(2009)}]{namer2009.hermes}
Namer, Fiammetta. 2009.
\newblock \emph{Morphologie, lexique et traitement automatique des langues :
  L'analyseur dérif}.
\newblock Paris: Hermès Science-Lavoisier.

\bibitem[{Namer \& Hathout(2025)}]{namer2025.elsevier}
Namer, Fiammetta \& Nabil Hathout. 2025.
\newblock Paradigms in morphology.
\newblock In \emph{International encyclopedia of language and linguistics, 3rd
  edition}, Elsevier.
\newblock Https://doi.org/10.1016/B978-0-323-95504-1.00505-6.

\bibitem[{Namer \& Hathout(to appear)}]{namer2026.compendium}
Namer, Fiammetta \& Nabil Hathout. to appear.
\newblock Word formation paradigms in the 21st century: does derivation mirror
  inflection, the reverse, or neither?
\newblock In Alexandra Bagasheva \& Jesús Fernández-Domínguez (eds.),
  \emph{21st century word-formation - a compendium}, De Gruyter.

\bibitem[{Namer et~al.(2023)Namer, Hathout, Amiot, Barque, Bonami, Boyé,
  Calderone, Cattini, Dal, Delaporte, Duboisdindien, Falaise, Grabar, Haas,
  Henry, Huguin, Juniarta, Liégeois, Lignon, Macchi, Manucharian, Masson,
  Montermini, Okinina, Sajous, Sanacore, Thi~Tran, Thuilier, Toussaint \&
  Tribout}]{namer2023.demonette}
Namer, Fiammetta, Nabil Hathout, Dany Amiot, Lucie Barque, Olivier Bonami,
  Gilles Boyé, Basilio Calderone, Julie Cattini, Georgette Dal, Alexander
  Delaporte, Guillaume Duboisdindien, Achille Falaise, Natalia Grabar, Pauline
  Haas, Frédérique Henry, Mathilde Huguin, Nyoman Juniarta, Loïc Liégeois,
  Stéphanie Lignon, Lucie Macchi, Grigoriy Manucharian, Caroline Masson, Fabio
  Montermini, Nadejda Okinina, Franck Sajous, Daniele Sanacore, Mai Thi~Tran,
  Juliette Thuilier, Yannick Toussaint \& Delphine Tribout. 2023.
\newblock Démonette-2, a derivational database for {F}rench with broad lexical
  coverage and fine-grained morphological descriptions.
\newblock \emph{Lexique} 33. 6--40.

\bibitem[{Plag(2003)}]{plag2003.word-formation-english}
Plag, Ingo. 2003.
\newblock \emph{Word-formation in english}.
\newblock Cambridge: Cambridge University Press.

\bibitem[{Plag \& Baayen(2009)}]{plag2009.suffix_ordering}
Plag, Ingo \& Harald Baayen. 2009.
\newblock Suffix ordering and morphological processing.
\newblock \emph{Language} 85(1). 109--152.

\bibitem[{Plénat(2005)}]{plenat2005.breves-remarques-ette}
Plénat, Marc. 2005.
\newblock Brèves remarques sur les déverbaux en \mbox{\emph{-ette}}.
\newblock In Frédéric Lambert \& Henning Nølke (eds.), \emph{La syntaxe au
  coeur de la grammaire. recueil offert en hommage pour le 60\ieme\
  anniversaire de claude muller}, 245--258. Rennes: Presses universitaires de
  Rennes.

\bibitem[{Plénat \& Roché(2014)}]{plenat2014.foisonnement-at}
Plénat, Marc \& Michel Roché. 2014.
\newblock La suffixation dénominale en \emph{\mbox{-at}} et la loi des (sous-)
  séries.
\newblock In Florence Villoing, Sophie David \& Sarah Leroy (eds.),
  \emph{Foisonnements morphologiques. Études en hommage à françoise
  kerleroux.}, 47--74. Nanterre: Presses Universitaires de Paris Ouest.

\bibitem[{Roch{\'e}(2011)}]{roche2011.linguistica}
Roch{\'e}, Michel. 2011.
\newblock Pression lexicale et contraintes phonologiques dans la d{\'e}rivation
  en \emph{\mbox{-aie}} du fran{\c c}ais.
\newblock \emph{Linguistica} 51. 299--315.

\bibitem[{Roché(2004)}]{roche2004.verbum}
Roché, Michel. 2004.
\newblock Mot construit ? {M}ots non construits ? {Q}uelques réflexions à
  partir des dérivés en \mbox{\emph{-ier(e)}}.
\newblock \emph{Verbum} 26. 459--480.

\bibitem[{Roché(2011)}]{roche2011.isme-dumal}
Roché, Michel. 2011.
\newblock Quel traitement unifié pour les dérivations en \mbox{\emph{-isme}}
  et en \mbox{\emph{-iste}} ?
\newblock In Michel Roché, Gilles Boyé, Nabil Hathout, Stéphanie Lignon \&
  Marc Plénat (eds.), \emph{Des unités morphologiques au lexique}, 69--143.
  Hermès Science-Lavoisier.

\bibitem[{Roché \& Plénat(2016)}]{roche2016.cmlf}
Roché, Michel \& Marc Plénat. 2016.
\newblock De l'harmonie dans la construction des mots français.
\newblock In \emph{Actes du 5\ieme\ congrès mondial de linguistique française
  ({CMLF} 2016)}, 1863--1878.

\bibitem[{Sajous et~al.(2020)Sajous, Calderone \&
  Hathout}]{sajous2020.lrec-englawi}
Sajous, Franck, Basilio Calderone \& Nabil Hathout. 2020.
\newblock {ENGLAWI}: From human{-} to machine-readable {W}iktionary.
\newblock In \emph{Proceedings of the twelfth international conference on
  language resources and evaluation ({LREC} 2020)}, 3009--3019. Marseille.

\bibitem[{Sajous \& Hathout(2015)}]{sajous2015.glawi}
Sajous, Franck \& Nabil Hathout. 2015.
\newblock {GLAWI}, a free {XML}-encoded {M}achine-{R}eadable {D}ictionary built
  from the {F}rench {W}iktionary.
\newblock In \emph{Proceedings of the of the {eLex} 2015 conference}, 405--426.
  Herstmonceux, England.

\bibitem[{Salvadori \& Huyghe(2023)}]{salvadori2023.polyfunctional}
Salvadori, Justine \& Richard Huyghe. 2023.
\newblock Affix polyfunctionality in {F}rench deverbal nominalizations.
\newblock \emph{Morphology} 33(1). 1--39.
\newblock \doi{10.1007/s11525-022-09401-4}.

\bibitem[{{\v{S}}ev{\v{c}}{\'i}kov{\'a}(2025)}]{sevcikova2025.habilitation}
{\v{S}}ev{\v{c}}{\'i}kov{\'a}, Magda. 2025.
\newblock \emph{A paradigmatic account of word formation: Conversion between
  noun and verb in {C}zech}.
\newblock Prague: Faculty of Arts, Charles University Habilitation.

\bibitem[{{\v{S}}ev{\v{c}}{\'i}kov{\'a} \&
  Kyjánek(2019)}]{sevcikova2019.semantics-derinet}
{\v{S}}ev{\v{c}}{\'i}kov{\'a}, Magda \& Lukáš Kyjánek. 2019.
\newblock Introducing semantic labels into the {D}eri{N}et network.
\newblock \emph{Journal of Linguistics/Jazykovedný casopis} 70(2). 412–423.
\newblock \doi{10.2478/jazcas-2019-0070}.

\bibitem[{{\v{S}}ev{\v{c}}{\'i}kov{\'a} \&
  {\v{Z}}abokrtsk{\'y}(2014)}]{sevcikova2014.LREC}
{\v{S}}ev{\v{c}}{\'i}kov{\'a}, Magda \& Zden{\v{e}}k {\v{Z}}abokrtsk{\'y}.
  2014.
\newblock Word-formation network for {C}zech.
\newblock In \emph{Proceedings of the ninth international conference on
  language resources and evaluation ({LREC}'14)}, 1087--1093. Reykjavik,
  Iceland.

\bibitem[{Silfverberg \& Hulden(2018)}]{silfverberg2018.pcfp}
Silfverberg, Miikka \& Mans Hulden. 2018.
\newblock An encoder-decoder approach to the paradigm cell filling problem.
\newblock In Ellen Riloff, David Chiang, Julia Hockenmaier \& Jun{'}ichi Tsujii
  (eds.), \emph{Proceedings of the 2018 conference on empirical methods in
  natural language processing}, 2883--2889. Brussels, Belgium: Association for
  Computational Linguistics.

\bibitem[{Strnadová(2014)}]{strnadova2014.phd}
Strnadová, Jana. 2014.
\newblock \emph{Les réseaux adjectivaux : sur la grammaire des adjectifs
  dénominaux en français}: Université Paris Diderot / Univerzita Karlova V
  Praze Thèse de doctorat.

\bibitem[{Stroppa(2005)}]{stroppa2005.these}
Stroppa, Nicolas. 2005.
\newblock \emph{Définitions et caractérisations de modèles à base
  d'analogies pour l'apprentissage automatique des langues naturelles}.
\newblock Paris: École nationale supérieure des télécommunications Thèse
  de doctorat.

\bibitem[{Stroppa \& Yvon(2005)}]{stroppa2005.conll}
Stroppa, Nicolas \& François Yvon. 2005.
\newblock An analogical learner for morphological analysis.
\newblock In \emph{Proceedings of the 9th conference on computational natural
  language learning (conll-2005)}, 120--127. Ann Arbor, MI: ACL.

\bibitem[{Talamo et~al.(2016)Talamo, Celata \&
  Bertinetto}]{talamo2016.derivatario}
Talamo, Luigi, Chiara Celata \& Pier~Marco Bertinetto. 2016.
\newblock {DerIvaTario}: {A}n annotated lexicon of {I}talian derivatives.
\newblock \emph{Word Structure} 9(1). 72--102.

\bibitem[{Tanguy \& Hathout(2002)}]{tanguy2002.taln}
Tanguy, Ludovic \& Nabil Hathout. 2002.
\newblock Webaffix : {U}n outil d'acquisition morphologique dérivationnelle à
  partir du {W}eb.
\newblock In Jean-Marie Pierrel (ed.), \emph{Actes de la 9\ieme\ conférence
  annuelle sur le traitement automatique des langues naturelles (taln-2002)},
  245--254. Nancy: ATALA.

\bibitem[{Thornton(2012)}]{thornton2012.overabundance}
Thornton, Anna~M. 2012.
\newblock Reduction and maintenance of overabundance. a case study on {I}talian
  verb paradigms.
\newblock \emph{Word Structure} 5(2). 183--207.

\bibitem[{Thym{\'e}-Gobbel et~al.(1994)Thym{\'e}-Gobbel, Ackerman \&
  Elman}]{thyme1994finnish}
Thym{\'e}-Gobbel, Ann, Farrell Ackerman \& Jeffrey Elman. 1994.
\newblock {F}innish nominal inflection: Paradigmatic patterns and token
  analogy.--the reality of linguistic rules.
\newblock In Susan~D Lima, Gregory Iverson \& Roberta Corrigan (eds.),
  \emph{The reality of linguistic rules}, John Benjamins Publishing Company.

\bibitem[{Tribout(2010)}]{tribout2010.phd}
Tribout, Delphine. 2010.
\newblock \emph{Les conversions de nom à verbe et de verbe à nom en
  français}: Université Paris 7 Thèse de doctorat.

\bibitem[{Tribout(2012)}]{tribout2012.word-structure}
Tribout, Delphine. 2012.
\newblock Verbal stem space and verb to noun conversion in {F}rench.
\newblock \emph{Word Structure} 5(1). 109--128.

\bibitem[{Vidra et~al.(2019)Vidra, {\v Z}abokrtsk{\'y}, Kyj{\'a}nek, {\v
  S}ev{\v c}{\'{\i}}kov{\'a} \& Dohnalov{\'a}}]{derinet2019.database}
Vidra, Jon{\'a}{\v s}, Zden{\v e}k {\v Z}abokrtsk{\'y}, Luk{\'a}{\v s}
  Kyj{\'a}nek, Magda {\v S}ev{\v c}{\'{\i}}kov{\'a} \& {\v S}{\'a}rka
  Dohnalov{\'a}. 2019.
\newblock {DeriNet} 2.0.
\newblock {LINDAT}/{CLARIN} digital library at the Institute of Formal and
  Applied Linguistics ({{\'U}FAL}), Faculty of Mathematics and Physics, Charles
  University.
\newblock \urlprefix\url{http://hdl.handle.net/11234/1-2995}.

\bibitem[{Vodolazsky(2020)}]{vodolazsky2020.derivbase-RU}
Vodolazsky, Daniil. 2020.
\newblock {D}eriv{B}ase.{R}u: a derivational morphology resource for {R}ussian.
\newblock In \emph{Proceedings of the twelfth language resources and evaluation
  conference}, 3937--3943. Marseille, France.

\bibitem[{Wauquier et~al.(2020)Wauquier, Fabre \& Hathout}]{wauquier2020.zwjw}
Wauquier, Marine, Cécile Fabre \& Nabil Hathout. 2020.
\newblock Semantic discrimination of technicality in {F}rench nominalizations.
\newblock \emph{Zeitschrift für Wortbildung / Journal of Word Formation}
  2020(2). 100--121.

\bibitem[{Williams et~al.(2020)Williams, Pimentel, Blix, McCarthy, Chodroff \&
  Cotterell}]{williams2020.form-meaning}
Williams, Adina, Tiago Pimentel, Hagen Blix, Arya~D. McCarthy, Eleanor Chodroff
  \& Ryan Cotterell. 2020.
\newblock Predicting declension class from form and meaning.
\newblock In Dan Jurafsky, Joyce Chai, Natalie Schluter \& Joel Tetreault
  (eds.), \emph{Proceedings of the 58th annual meeting of the association for
  computational linguistics}, 6682--6695. Online: Association for Computational
  Linguistics.

\bibitem[{Ylonen(2022)}]{ylonen2022.lrec}
Ylonen, Tatu. 2022.
\newblock Wiktextract: Wiktionary as machine-readable structured data.
\newblock In \emph{Proceedings of the language resources and evaluation
  conference}, 1317--1325. Marseille, France.

\bibitem[{Zeller et~al.(2014)Zeller, Padó \& Šnajder}]{zeller2014.coling}
Zeller, Britta~D., Sebastian Padó \& Jan Šnajder. 2014.
\newblock Towards semantic validation of a derivational lexicon.
\newblock In \emph{Proceedings of {COLING}}, 1728--1739. Dublin, Ireland.

\bibitem[{Zeller et~al.(2013)Zeller, Šnajder \& Padó}]{zeller2013.derivbase}
Zeller, Britta~D., Jan Šnajder \& Sebastian Padó. 2013.
\newblock D{ErivBase}: {I}nducing and evaluating a derivational morphology
  resource for {G}erman.
\newblock In \emph{Proceedings of acl}, 1201--1211. Sofia, Bulgaria.

\end{thebibliography}
\end{document}